\documentclass[10pt,twocolumn,letterpaper]{article}

\usepackage{cvpr}              

\usepackage{graphicx}
\usepackage{amsmath}
\usepackage{amssymb}
\usepackage{booktabs}
\usepackage{cuted}

\usepackage[pagebackref,breaklinks,colorlinks]{hyperref}

\usepackage[capitalize]{cleveref}
\crefname{section}{Sec.}{Secs.}
\Crefname{section}{Section}{Sections}
\Crefname{table}{Table}{Tables}
\crefname{table}{Tab.}{Tabs.}

\def\cvprPaperID{6562} 
\def\confName{CVPR}
\def\confYear{2022}

\begin{document}

\title{Pixel-Level Face Image Quality Assessment for Explainable Face Recognition}

\newcommand\Mark[1]{\textsuperscript#1}

\author{Philipp Terh\"{o}rst\Mark{1}\Mark{2}, Marco Huber\Mark{2}\Mark{3}, Naser Damer\Mark{2}, Florian Kirchbuchner\Mark{2}\Mark{3}, Kiran Raja\Mark{1}, Arjan Kuijper\Mark{2}\Mark{3}\\
\Mark{1}Norwegian University of Science and Technology, Gj{\o}vik, Norway\\ 
\Mark{2}Fraunhofer Institute for Computer Graphics Research IGD, Darmstadt, Germany\\
\Mark{3}Technical University of Darmstadt, Darmstadt, Germany\\
Email:{\{philipp.terhoerst, marco.huber, naser.damer, florian.kirchbuchner, arjan.kuijper\}@igd.fraunhofer.de}
}


\maketitle

\begin{strip}
\centering
\includegraphics[width=1.0\textwidth]{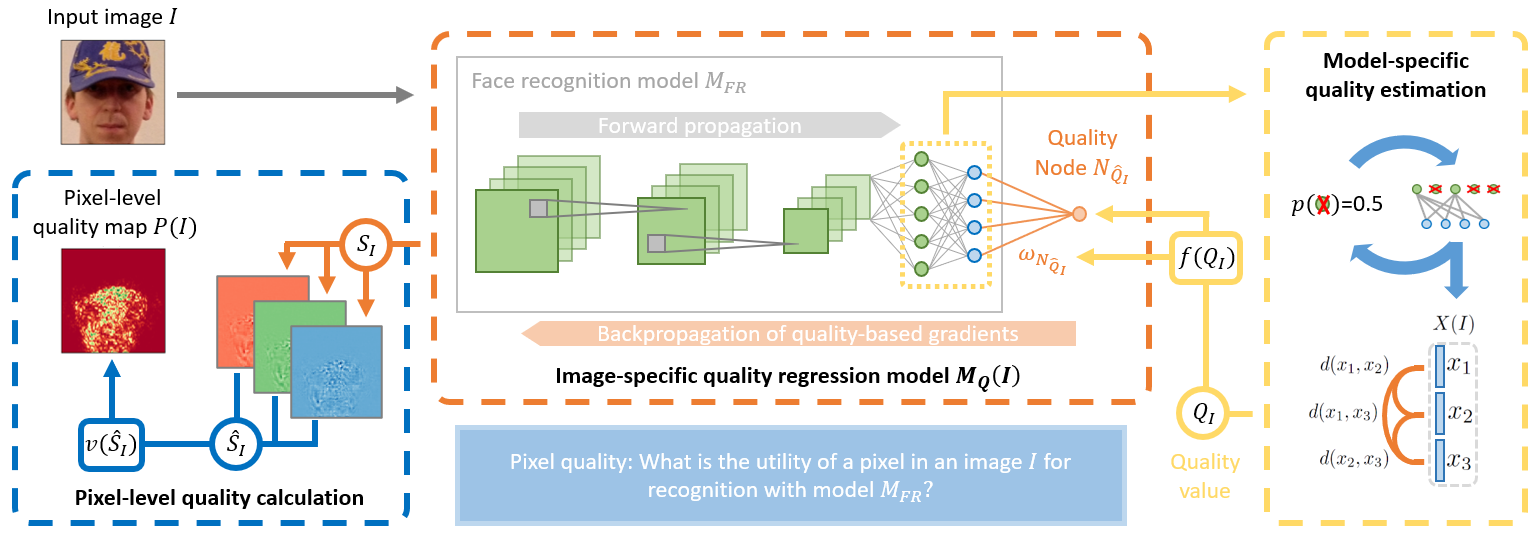}
\captionof{figure}{\textbf{Overview of the proposed pixel-level quality estimation approach} in three steps. First, the input image $I$ is passed through a face recognition model $\mathcal{M}_{FR}$ with repetitive forward-passes through the last layer to obtain a model-specific quality estimate $Q_I$. This quality value is modified and used in the second step to build an image-specific quality regression model $\mathcal{M_Q}(I)$ by extending $\mathcal{M}_{FR}$ with a quality node $N_{\hat{Q}_I}$ that is connected through the weights $\omega_{N_{\hat{Q}_I}}$. In the third step, the constructed regression model $\mathcal{M_Q}(I)$ is used to backpropagate the quality-based gradients and to transform the resulting saliency maps to a pixel-level quality map $P(I)$. }
\label{fig:Overview}
\end{strip}

\begin{abstract}
An essential factor to achieve high performance in face recognition systems is the quality of its samples.
Since these systems are involved in daily life there is a strong need of making face recognition processes understandable for humans.
In this work, we introduce the concept of pixel-level face image quality that determines the utility of pixels in a face image for recognition.
We propose a training-free approach to assess the pixel-level qualities of a face image given an arbitrary face recognition network.
To achieve this, a model-specific quality value of the input image is estimated and used to build a sample-specific quality regression model.
Based on this model, quality-based gradients are back-propagated and converted into pixel-level quality estimates.
In the experiments, we qualitatively and quantitatively investigated the meaningfulness of our proposed pixel-level qualities based on real and artificial disturbances and by comparing the explanation maps on faces incompliant with the ICAO standards.
In all scenarios, the results demonstrate that the proposed solution produces meaningful pixel-level qualities enhancing the interpretability of the complete face image quality. The code is publicly available\footnote{\url{https://github.com/pterhoer/ExplainableFaceImageQuality}}.

\end{abstract}

\vspace{-5mm}
\section{Introduction}
\label{sec:Introduction}

Face recognition (FR) systems are spreading worldwide and have a growing effect on our daily lifes \cite{DBLP:journals/ijon/WangD21a}.
Since these systems are increasingly involved in critical decision-making processes, such as in forensics and law enforcement, there is a growing need in making the FR process explainable to humans \cite{DBLP:journals/corr/abs-2009-01103}.
Especially in unconstrained environments, FR systems have to deal with large variabilities, such as image acquisition conditions (illumination, background) and factors of the face (pose, occlusions, expressions), that might result in defective matching decisions \cite{ISO19794-5-2011, ICAO9303}.
The impact of these variabilities on the FR performance is measured in terms of face image quality (FIQ).
Consequently, the performance of FR systems is strongly dependent on the quality of their samples. 
The FIQ of a sample is defined as its utility for recognition \cite{DBLP:journals/corr/abs-1904-01740, 6712715, 10.1007/978-3-540-74549-5_26, DBLP:journals/corr/Best-RowdenJ17}.
The automatic prediction of FIQ (prior to matching) is one of the key factors during the enrolment and is essential to achieve robust and accurate FR performances \cite{DBLP:journals/corr/abs-2009-01103}.

Previous research on FIQ focused mainly on the development of accurate quality assessment methods \cite{DBLP:conf/cvpr/TerhorstKDKK20, 6877651, DBLP:journals/corr/Best-RowdenJ17, Ou_2021_CVPR, DBLP:journals/corr/abs-1904-01740}.
Although these methods possess similar bias problems than for FR systems \cite{DBLP:conf/icb/TerhorstKDKK20}, no works aimed at making the output of FIQ assessment (FIQA) methods explainable to humans \cite{DBLP:journals/corr/abs-2009-01103}, and thus provide an interpretable reason for a face image being of low or high quality.
On the other hand, previous works on explainable FR focused solely on making the matching decision explainable to humans, neglecting that explainability is also needed during the enrolment of subjects, where the quality compliance of the image is typically checked.

In this work, we propose a training-free approach to compute pixel-level quality (PLQ) explanation maps that determines the utility of single pixels for recognition, similar to the definition of FIQ.
The PLQ-maps aim at making the enrolment process explainable for humans.
Their construction consists of three steps as shown in Figure \ref{fig:Overview}.
First, a model-specific quality value for an input image is estimated.
Second, this quality value and the FR model are used to build a sample-specific quality regression model without the need for training.
Third, this quality model, optimized for the input image, is used to back-propagate quality-based gradients and convert these into PLQ estimates.

In the experiments, the effectiveness of the proposed PLQ-maps are evaluated quantitatively and qualitatively in three scenarios.
This is done by demonstrating that areas of low pixel-quality result in lower FIQ values and vice versa.
First, it is shown that inpainting low pixel-quality areas in the face (such as occlusions) localised by our method increases the FIQ.
Second, it is demonstrated that placing random disturbances on the face results in easily-detectable areas of low pixel-quality.
Third, the PLQ-maps are analysed based on face images incompliant to various International Civil Aviation Organization (ICAO) specifications  \cite{ICAO9303}.
In all three scenarios, the results demonstrate that the proposed solution produces meaningful PLQ values.

The proposed PLQ-maps have several advantages.
First, they can be used to deepen the understanding of how FIQA and FR models work since these maps describes the importance of pixels for the model-specific FIQ and therefore also for the FR model.
Second, they can be used to enhance FR performance by inpainting or merging low-quality areas of the face to create images of higher utility.
Last, they can explain why an image cannot be used as a reference image during the acquisition/enrolment process and in which area of the face the subject have to do changes to increase the quality.
Consequently, PLQ maps provide guidance on the reasons behind low quality images, and thus can provide interpretable instructions to improve the FIQ.

To summarize, the proposed PLQA approach (a) can be applied on arbitrary FR networks, (b) does not require training, and (c) provides a pixel-level utility description of an input face explaining how well pixels in face image are suited for recognition before it is  used for matching.

\section{Related Work}
\label{sec:RelatedWork}

\subsection{Explainable Face Recognition}

Explainable FR is a relatively new field of research that aims at making the face recognition pipeline, and its consequences, explainable for humans.
In 2019, Yin et al. \cite{DBLP:conf/iccv/YinTLS019} proposed a spatial activation diversity loss.
The loss penalizes correlations among filter weights and they showed that their filter distribution is more spread to different spatial areas.
This lead to learned face representations of higher structure and therefore higher interpretability since each dimension of the representation represents a face structure or a face part.
In \cite{DBLP:conf/iccvw/ZeeGN19}, Zee et al.\ trained a classification network on faces and used class activation maps to find the most distinguishable regions.
With this information, the authors showed that the human FR performance is increased.
In 2020, Williford et al. \cite{DBLP:conf/eccv/WillifordMB20} proposed new approaches for explainable FR.
Based on triplets consisting of a probe, a mate, and a non-mate image, the algorithms generate saliency maps that highlight the maximum similarity between the probe and the mate and the minimum between the probe and the non-mate.
This  provides  explanations  on  why  the  matcher comes to a certain decision.

So far, works on explainable FR have focused on making the matching decision explainable. 
Contrarily, in this work, we propose a method to make the utility of an image for recognition explainable before any matchings.





\subsection{Face Image Quality Assessment}

Several standards have been proposed to ensure face image quality by constraining the capture requirements, such as ISO/IEC 19794-5 \cite{ISO19794-5-2011} and ICAO 9303 \cite{ICAO9303}.
These standards divide quality into \textit{image-based} qualities  (such as illumination, occlusion) and \textit{subject-based} quality measures (such as pose, expression, accessories).
This influenced the first generation of FIQA approaches that are built on human perceptive image quality factors \cite{10.1007/978-3-540-74549-5_26, 6197711, 7935089, 5424029, 6712715, 6985846, 4341617, 6460821, 6996248}.
However, due to the achieved performance, the research focus shifted to learning-based approaches.
%
The second generation of FIQA approaches \cite{5981881, 5981784, 6877651, 7351562, DBLP:journals/corr/Best-RowdenJ17, DBLP:journals/corr/abs-1904-01740} consists of supervised learning algorithms based on human or artificially constructed quality labels.
These quality labels were either based on human judgement or derived from comparison score distributions.
The utilized algorithms include rank-based learning \cite{6877651}, the use of SVM-based approaches \cite{DBLP:journals/corr/Best-RowdenJ17}, and training deep networks with artificial quality labels \cite{DBLP:journals/corr/abs-1904-01740, Ou_2021_CVPR, DBLP:journals/corr/abs-2009-00603}.
However, humans may not know the best characteristics for face recognition systems and artificially labelled quality values, derived from comparison scores, rely on error-prone labelling mechanisms and require large-scale training.

The third generation of FIQA approaches completely avoids the use of quality labels.
In 2020, Terh\"{o}rst et al. \cite{DBLP:conf/cvpr/TerhorstKDKK20} proposed stochastic embedding robustness for FIQA (SER-FIQ).
This concept measures the robustness of a face representation against dropout variations and uses this measure to determine the quality of a face.
It avoids the need for training and takes into account the decision patterns of the deployed face recognition model.
In 2021, Meng et al. \cite{Meng_2021_CVPR} proposed a class of loss functions that include magnitude-aware angular margins, encoding the quality into the face representation.
Training with this loss results in an FR model that produces embeddings whose magnitudes can measure the FIQ of their faces.

So far, research on FIQ focuses only on the development of FIQA methods.
Although that it was shown that FIQA possesses similar bias problems than for FR \cite{DBLP:conf/icb/TerhorstKDKK20}, no works aimed at making the output of FIQA explainable to humans.
While there are various approaches to visualize classification decisions of deep learning models \cite{DBLP:journals/ijcv/SelvarajuCDVPB20,DBLP:journals/corr/SpringenbergDBR14, DBLP:journals/tip/MopuriGB19}, to the best of our knowledge, this is the first work on explaining the utility of face representations.


\section{Methodology}
\label{sec:Methodology}

The proposed pixel-level quality estimation method consists of three steps.
First, for a given face image $I$, a model-specific quality estimate is computed stating its utility for the face recognition network.
Second, the quality value and the recognition network are used to build a quality regression model without the need for training.
Third, this model is used to back-propagate quality-based gradients and convert these into pixel-level face image quality estimations.
An overview of the proposed concept is shown in Figure \ref{fig:Overview}.


\subsection{Model-Specific Quality Estimation}
\label{sec:SERFIQ}

To compute the model-specific FIQ value $Q_I$, our method builds on the work of Terh\"orst et al. \cite{DBLP:conf/cvpr/TerhorstKDKK20}. 
This choice is based on its training-free applicability to arbitrary FR networks and since it determines how well a specific model $\mathcal{M}_{FR}$ can use $I$ for recognition.
Given a face image $I$, this image is propagated through the network and the forward passes to the last (embedding) layer are repeated $m=100$ times as motivated in \cite{DBLP:conf/cvpr/TerhorstKDKK20}. 
During each of these stochastic forward passes, a different dropout pattern (with $p_d=0.5$) is applied resulting in a set of $m$ different stochastic embeddings $X_I$.
The FIQ of the image $I$ is given by
\begin{align}
Q_I = Q(X_I) = 2 \sigma \left( -\dfrac{2}{m^2} \textstyle \sum_{i<j} d(x_i,x_j)  \right) ,
\end{align}
where $\sigma(\cdot)$ is the sigmoid function and states the Euclidean distance between two stochastic embedding $x_i,x_j \in X_I$.
$Q_I$ defines the quality of an image over the robustness of its embeddings.
If there are high variations in the stochastic embeddings, the robustness of the representation is low and thus the quality.
Since the quality score is model-dependent and often in a narrow range, we additionally adjust the score to the range of $[0,1]$ using
\begin{align}
\hat{Q}_I = f(Q_I) = \sigma \left(\alpha (Q_I - r) \right). \label{eq:QualityScaling}
\end{align}
Choosing $r$ near the mean of the quality distribution of a development set ensures a new mean quality around 0.5 after applying Eq. \ref{eq:QualityScaling}.
Parameter $\alpha$ is chosen to stretch the values to a range of $[0,1]$. 
Please note that this quality scaling is optional and only aims at making the results more easily comparable.

\subsection{Building a Quality Regression Model}
\label{sec:BuildingQualityRegressionModel}

Based on this face image quality score $\hat{Q}_I$ and the face recognition model $\mathcal{M}_{FR}$, we now build a quality regression model $\mathcal{M}_Q$ in a training-free fashion.
This is performed by extending the face recognition model $\mathcal{M}_{FR}$ with a one-dimensional quality node $N_{\hat{Q}_I}$.
The node is fully connected to the (last) embedding layer of $\mathcal{M}_{FR}$.
The weights of these connections are given by
\begin{align}
w_{N_{\hat{Q}_I}} = \textstyle \dfrac{\hat{Q}_I}{||e_I||_1} ,
\end{align}
where $\mathcal{M}_{FR}(I) = e_I$ is the face embedding of $\mathcal{M}_{FR}$ for $I$.
This assumes linear layer activation with a bias term of $b=0$ and ensures that all features of $e_I$ are equally important for the quality estimation.
Moreover, the construction of $\mathcal{M}_Q$ ensures that given $I$, the output of the model is $\hat{Q}_I$.

\subsection{Pixel-Level Quality Calculation}
\label{sec:PixelQuality}

%

The constructed quality regression model $\mathcal{M}_Q$ for image $I$ is, similarly to $\mathcal{M}_{FR}$, pairwise differentiable.
Therefore, we can compute a gradient-based saliency map
\begin{align}
S(I) = \dfrac{\delta \mathcal{M}_Q(I) }{\delta I} ,
\end{align}
similar to \cite{DBLP:journals/corr/SimonyanVZ13, DBLP:journals/jmlr/BaehrensSHKHM10, DBLP:journals/corr/SmilkovTKVW17}.
In contrast to these, the saliency map in our work is not dependent on a certain class but rather on the continuous quality value.
The saliency map $S(I)$ consists of the gradients for each pixel of $I$. 
The magnitudes of these gradients indicate the relative effect on each pixel on the FIQ value.

Considering $S(I)$, only the magnitudes of the gradients are crucial for the quality assessment task while their directions are context-dependent \cite{DBLP:journals/corr/SmilkovTKVW17}.
Consequently, the three color channels of $S(I)$ are merged considering only the absolute values of the gradients.
This is done by 
\begin{align}
\hat{S}(I) = \dfrac{1}{3} \sum_{c=1}^3 |g_{i,j,c}| ,
\end{align}
where $g_{i,j,c}$ represents the gradient for pixel $(i,j)$ of color channel $c$.
$\hat{S}(I)$ can already be interpreted as pixel-level qualities.
However, since the pixel-level qualities aim at visually explaining the utility of an image for recognition in a human-understandable manner and the ranges of $\hat{S}(I)$ are, depending on $\mathcal{M}_{FR}$, in a narrow range, a visualization function $v$
\begin{align}
v(\hat{S}) = 1 - \dfrac{1}{1 + (10^{\gamma} \times \hat{g}_{i,j,c}^2)} , \label{eq:PixelQualityCalibration}
\end{align}
is used to project $\hat{S}$ to a more intuitive range of $[0,1]$.
The visualization parameter $\gamma$ is used to stretch the quality values to the desired range.
Applying $v(\cdot)$ on $\hat{S}(I)$ results in the pixel-level quality map
\begin{align}
P(I) = v\left(\hat{S}(I)\right) .
\end{align}
$P(I)$ is the representation of the pixel-level qualities $p_{i,j} \in [0,1]$.
A higher pixel quality indicates a higher contribution for the recognition utility of the face image and vice versa.

Please note that, in contrast to the typical procedure when dealing with gradient backpropagation, we do not scale the gradients per image, e.g. with MinMax scaling.
Scaling the gradients would highlight the differences in low- and high-quality regions of a single image but also result in the loss of the PLQ comparability between different images.
For instance, such scaling will always result in an area of high quality even if the image is not suitable for recognition.


%
%
%
%
%
%
%

\section{Experimental Setup}
\label{sec:ExperimentalSetup}

\subsection{Databases}
\label{sec:Databases}

\begin{figure}
\centering
\includegraphics[width=0.4\textwidth]{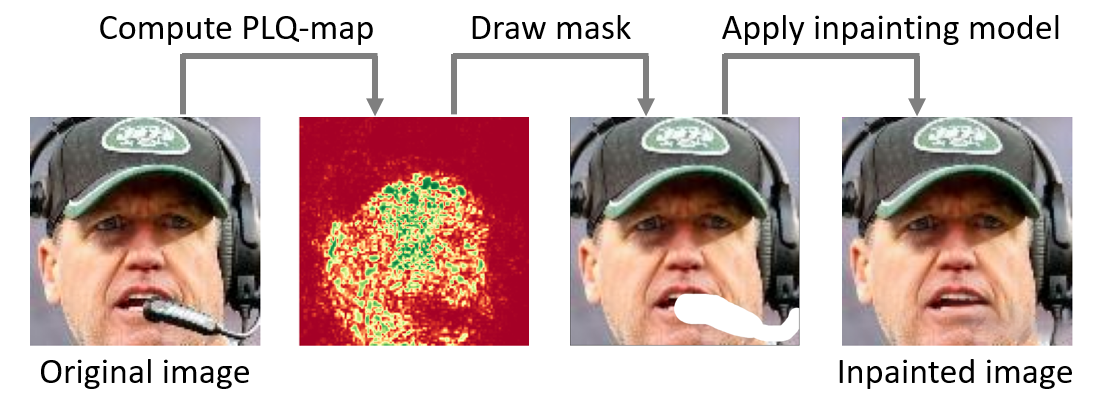}
\caption{\textbf{Workflow of the Pre- and Inpainted Dataset creation} - Using the proposed methodology, the biggest low-quality area in the face is located. Then, this area is masked, and inpainted \cite{10.1007/978-3-030-01252-6_6} to create a similar face image without the potential quality-decreasing factor. \vspace{-4mm}}
\label{fig:InpaintingDataset_Workflow}
\end{figure}

\paragraph{The Pre- and Inpainted dataset} was created by manually selecting images from VGGFace2 \cite{DBLP:conf/fgr/CaoSXPZ18} and Adience \cite{Eidinger:2014:AGE:2771306.2772049} since these contain images of large variances.
The decision criteria for the selection was that the images must contain occlusions or similar quality-decreasing factors according to human judgement.
Then, for each image, we applied the proposed method to identify the biggest low-pixel quality region and manually created an inpainting mask of this region.
The image and the corresponding mask are given to an inpainting model \cite{10.1007/978-3-030-01252-6_6}.
If the generated image consists of artefacts, the image is discarded.
This resulted in pairs of similar face images with and without quality-decreasing factors.
The workflow of the dataset creation is shown in Figure \ref{fig:InpaintingDataset_Workflow}.
The created dataset consists of 100 pairs of face images with their inpainted counterparts.
We will use this dataset to demonstrate the inpainted low-quality regions harm the face image quality by comparing the image qualities of the face pairs.

\paragraph{The Random Mask dataset} is based on the ColorFeret \cite{ColorFERET} database due to the high image-quality of its images which corresponds to scenarios such as identity document and border checks.
Each image of the random mask dataset was created by placing a black square on the inner image of a frontal face.
For each ColorFeret image, 5 black squares of size $s \times s$ ($s=10,20,30,40,50$) pixels were placed randomly on the image resulting in 5 images for the random mask dataset.
To avoid that non-facial areas are masked, the squares are only placed in the inner 90\% of the face images. 
This results in a total of 6610 masked face images to demonstrate that the proposed methodology can detect these disturbances as low-quality regions.

\paragraph{The Inhouse ICAO Incompliance dataset} was collected by us to analyse the effect of pixel-level face image quality on face images that violate various International Civil Aviation Organization (ICAO) specifications \cite{ICAO9303}.
It consists of a reference image of one subject that complies with these specifications as well as 33 face images of the same subject with different violations of these specifications.
The images were taken with fixed capturing conditions to allow a clear investigation of the effect of pixel-level face image qualities on ICAO incompliances.
In the supplementary material, we included a more detailed discussion, such as on the licenses and copyright.

\subsection{Face Recognition Models and Parameters}
\label{sec:FaceRecognitionModels}
We analysed the image- and pixel-level face quality based on two widely-used FR models using FaceNet\footnote{\url{https://github.com/davidsandberg/facenet}} \cite{DBLP:journals/corr/SchroffKP15} and ArcFace\footnote{\url{https://github.com/deepinsight/insightface}} \cite{Deng_2019_CVPR} losses (both MIT License) based on ResNet-100.
For the sake of simplicity, we refer to these models as FaceNet and ArcFace.
Both models were trained on the MS1M database \cite{DBLP:journals/corr/GuoZHHG16}.
Given a face image, the image is aligned, scaled, and cropped before being passed to one of the models.
This preprocessing is done as described in \cite{DBLP:journals/corr/abs-1812-01936} for ArcFace, and as described in \cite{Kazemi2014OneMF} for FaceNet.

Since the quality estimations are model-specific for FR systems, the parameters for the quality scaling and the quality visualization are as well.
The quality values are adjusted to a wider range of $[0,1]$ on the quality values of the Adience benchmark \cite{Eidinger:2014:AGE:2771306.2772049} and resulted in parameters $\alpha_{AF}=130, r_{AF}=0.88$ for ArcFace and $\alpha_{FN}=450, r_{FN}=0.93$ for FaceNet.
For visualizing the pixel-level qualities, we choose $\gamma_{AF} = 7.5$ for ArcFace and $\gamma_{FN} = 5.5$ for FaceNet.
Please note that the choice of $\gamma$ is subjective and depending on the colormap used for visualizing the quality values\footnote{We recommend to adjust $\gamma$ based on an ICAO compliant \cite{ICAO9303} face image such that the center of the face shows the high quality color (green) while the background shows a uniform color for low quality (red).}.
In general, the choice of these parameters ($\alpha, r, \gamma$) determine the scaling of the qualities and thus, aim to make the results more easily understandable for humans.
Since the scaling is done with a strictly increasing function, the order of the qualities, and thus the FIQA task in general, is not affected.


\subsection{Investigations}
\label{sec:Investigations}

The proposed pixel-level quality estimation approach is analysed from two directions.
First, low-quality face images with low-quality areas, such as occlusions, are localised and inpainted to demonstrate that this improves the FIQ.
Second, random masks are placed on high-quality faces to show that the proposed methodology identifies these as low-quality areas.
Both evaluation approaches, enhancing low-quality images and degrading high-quality images, aim at quantitatively (via quality-changes) and qualitatively (via changes in the PLQ-maps) investigating the effectiveness of the proposed PLQA approach.
Lastly, the PLQ-maps are investigated on ICAO-incompliant faces.

%

\section{Results and Discussions}
\label{sec:Results}


\subsection{Analysing PLQ-Change by Enhancing FIQ}
\label{sec:AnalysingPLQ_EnhancingFIQ}

Using the Pre- and Inpainted dataset presented in Section \ref{sec:Databases}, we examine the pixel-level quality explanation maps of images with occlusions and other quality degradations as well as images where these impairments have been corrected with inpainting.
Comparing the images before and after inpainting allows us to make a statement about how well the proposed solution works since the inpainted areas were determined as low pixel-quality areas by our method.

\begin{figure}
\centering
\subfloat[ArcFace \label{fig:QualityChangeInpainting_ArcFace}]{%
	\includegraphics[width=0.24\textwidth]{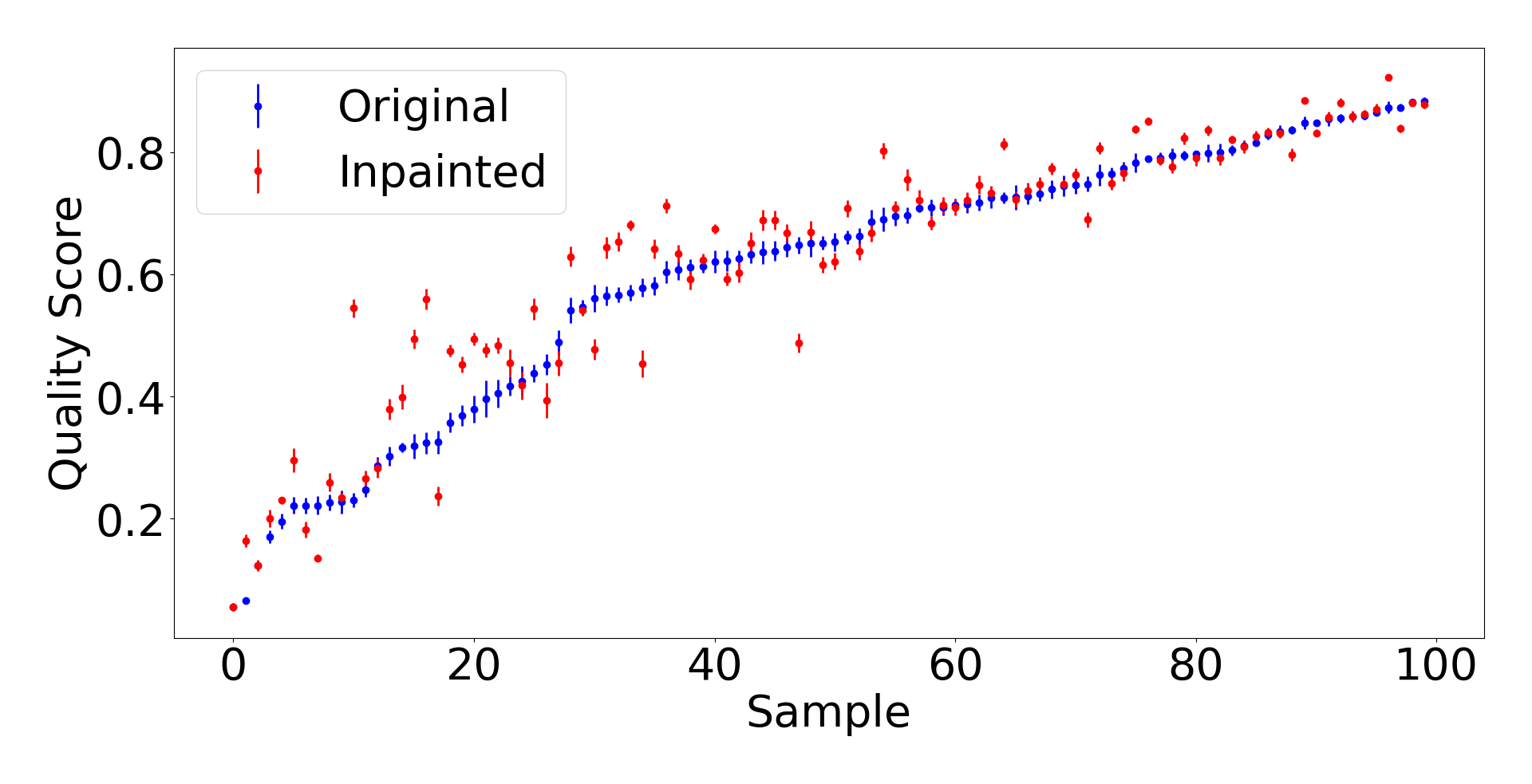}}  
\subfloat[FaceNet \label{fig:QualityChangeInpainting_FaceNet}]{%
	\includegraphics[width=0.24\textwidth]{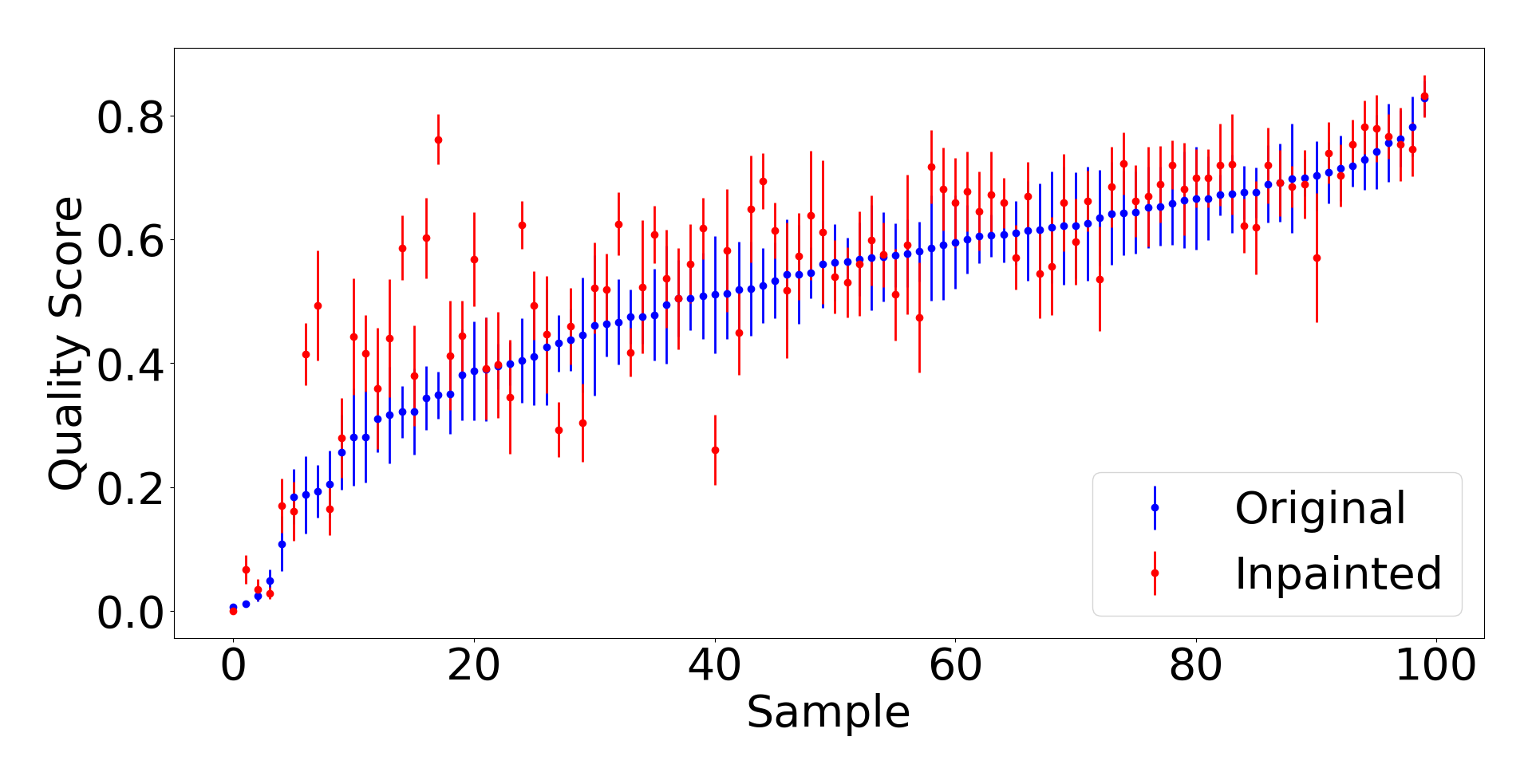}} 	
\caption{\textbf{Quality values before and after inpainting} - 
The values are sorted and plotted with their standard deviation (STD) of calculating the FIQ score 10 times. For ArcFace (FaceNet), the quality scores decreased in 15\% (5\%) of the cases while the inpainting improved the image quality in 65\% (69\%) of the cases.
In the remaining cases, the quality change was within the STD.
The decreases might origin from error-prone inpaintings. In general, inpainting low-quality areas improves the FIQ demonstrating that PLQ values are meaningful.
}
\label{fig:QualityChangeInpainting}
\end{figure}

\begin{figure*}
\centering
\includegraphics[width=0.95\textwidth]{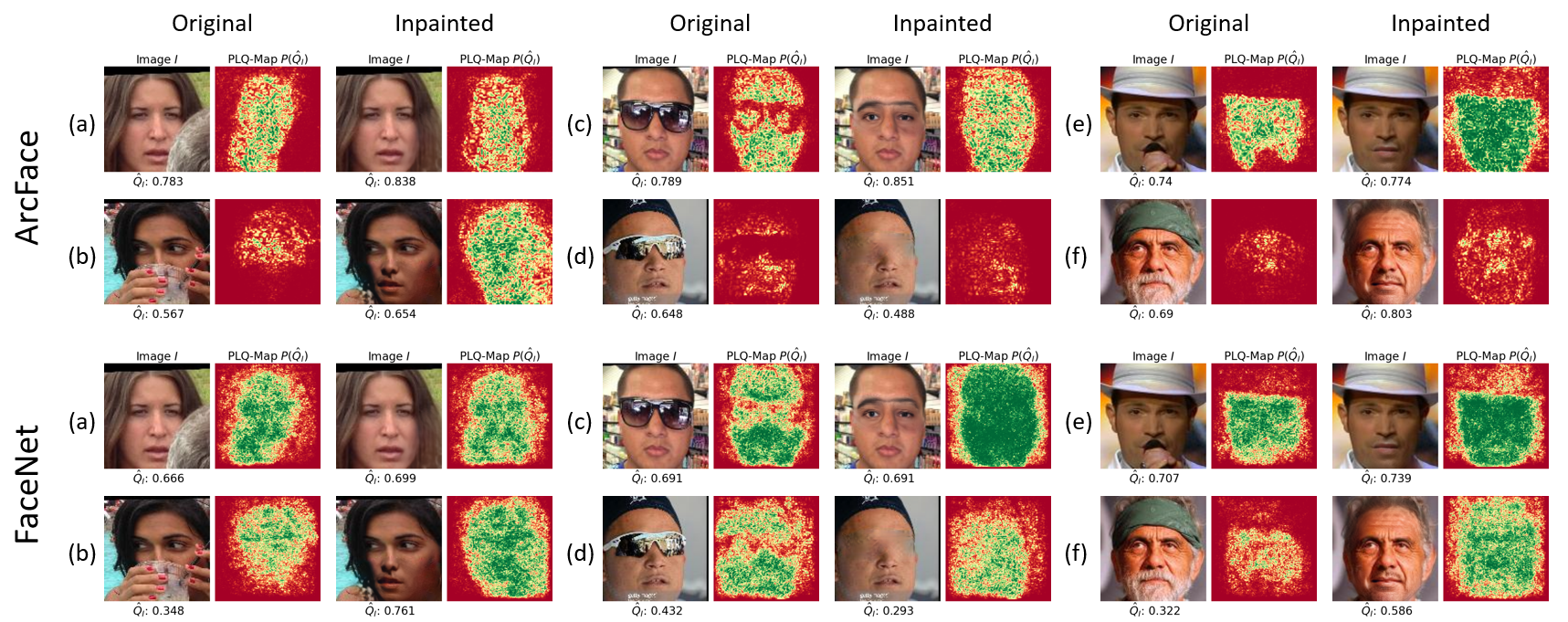}
\caption{\textbf{PLQ explanation maps before and after inpainting} - Images before and after the inpainting process are shown with their corresponding PLQ-maps and FIQ values. The images show the effect of small and large occlusions, glasses, headgears, and beards on the PLQ-maps for two FR models. In general, these are identified as areas of low pixel-quality and inpainting these areas strongly increases the pixel-qualities of these areas as well as the FIQ. This demonstrates that our solution leads to reasonable pixel-level quality estimates and thus can give interpretable recommendations on the causes of low quality estimates.}
\label{fig:InpaintingResults}
\end{figure*}

\vspace{-2mm}
\paragraph{Quantitative Analysis -}
Figure \ref{fig:QualityChangeInpainting} shows the FIQ values before and after the inpainting for two face recognition models.
For ArcFace, only 15\% of the quality scores decreased while 65\% of the scores increased with inpainting low pixel-quality areas.
For FaceNet, only 5\% of the inpaintings decreased the FIQ and in 69\%, the FIQ increased.
In the remaining cases, the quality change was negligible since it was within the standard deviation of the FIQ scores.
Moreover, many cases of decreased quality scores origin from bad inpaintings showing unreasonable artefacts (see Figure \ref{fig:InpaintingResults}(d)).
In general, inpainting areas that our method identified as low-quality improves the FIQ and thus, supports the validity of our approach.
Especially for images with low FIQ, inpaiting low pixel-quality areas lead to strong quality enhancements as shown in Figure \ref{fig:QualityChangeInpainting}.

\vspace{-2mm}
\paragraph{Qualitative Analysis -}
In Figure \ref{fig:InpaintingResults}, the PLQ explanation maps for two face recognition models are shown before and after the inpainting.
Figure \ref{fig:InpaintingResults}(a) shows a recovered cheek area of a face.
For both models, the PLQ-map of the original image shows low pixel-qualities in the area of the covered cheek.
In the inpainted image, the cheek is recovered and the area is determined as high-quality pixels.
In Figure \ref{fig:InpaintingResults}(b), a large occlusion covering the lower part of the face is shown. 
While for the PLQ-map on ArcFace this area is clearly detected, for FaceNet this is only shown as medium quality.
After inpainting, this area is recognized as high-quality by both models.
Figure \ref{fig:InpaintingResults}(c) shows the effect of glasses on the PLQ-maps.
For both models, the frame of the glasses is recognized as low quality and removing the glasses lead to high pixel-qualities.
Figure \ref{fig:InpaintingResults}(d) shows the case of a faulty inpainting.
Before the inpainting, the glasses and reflections are shown as low pixel-qualities.
After, the method failed to mark the missing eyes as low quality.
In Figure \ref{fig:InpaintingResults}(e) a small occlusion is shown.
For ArcFace, this occlusion is represented more sharply than for FaceNet.
However, removing this occlusion leads to high-pixel qualities for both models.
Moreover, the hat is sharply estimated as low-pixel quality for both models.
Lastly, Figure \ref{fig:InpaintingResults}(f) demonstrates the case of multiple occlusions (headgear and beard).
Both occlusions are marked as low-quality pixels and after the inpainting, the qualities are increased.
These examples demonstrate that the proposed solutions lead to reasonable pixel-level quality estimates.


\subsection{Analysing PLQ-Change by Decreasing FIQ}
\label{sec:AnalysingPLQ_DecreasingFIQ}

Using the Random Mask dataset described in Section \ref{sec:Databases}, we examine the pixel-level quality explanations by degrading the high-quality images using randomly placed masks.
If the PLQ-maps represent the mask areas as low-quality, we can conclude that our solutions can successfully detect such disturbances.

\begin{figure}
\centering
\subfloat[Image Quality Change - ArcFace \label{fig:ImageQualityChangeRandomMask_ArcFace}]{%
	\includegraphics[width=0.24\textwidth]{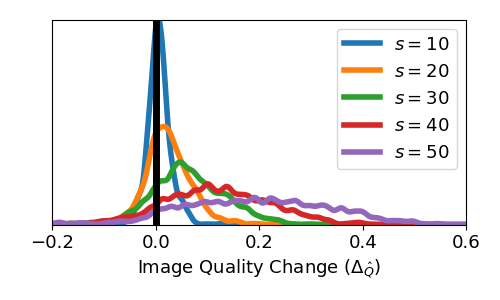}}  
\subfloat[Image Quality Change - FaceNet \label{fig:ImageQualityChangeRandomMask_FaceNet}]{%
	\includegraphics[width=0.24\textwidth]{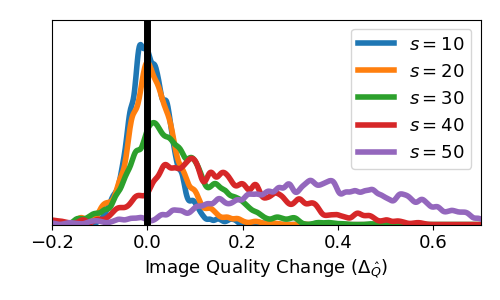}} 	
	
\subfloat[Pixel Quality Change - ArcFace \label{fig:PixelQualityChangeRandomMask_ArcFace}]{%
	\includegraphics[width=0.24\textwidth]{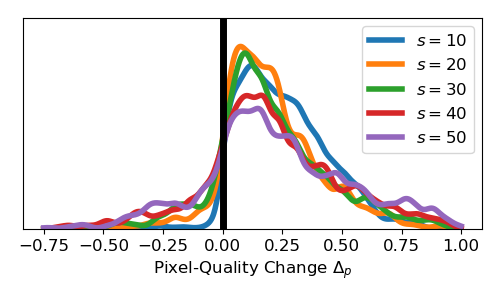}}  
\subfloat[Pixel Quality Change - FaceNet \label{fig:PixelQualityChangeRandomMask_FaceNet}]{%
	\includegraphics[width=0.24\textwidth]{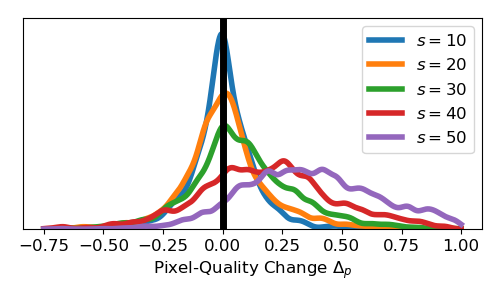}} 
\caption{\textbf{Quality changes through random masks} - 
High-quality images are degraded by placing random masks of size $s \times s$ pixels on the images. The effect of this is analysed in terms of FIQ change $\Delta_{\hat{Q}}$ of the image and in terms of mean pixel quality change $\Delta_p$ in the masked area. The distributions of the image quality changes for both models are shown in (a,b) and (c,d) present the distribution for the pixel quality changes.
Positive quality changes (values right of the black line) indicate that the disturbances degrade the qualities. Since the majority of the changes are positive, our solution is able to detect these disturbances and assigns them with low-qualities.
This holds especially true for larger disturbances for which FR systems are less robust at.\vspace{-4mm}
}
\label{fig:QualityChangeRandomMask}
\end{figure}

\vspace{-2mm}
\paragraph{Quantitative Analysis -}
Figure \ref{fig:QualityChangeRandomMask} shows the effect of the random masking process on the image- and pixel-level qualities for two face recognition models and five mask sizes.
In Figures \ref{fig:ImageQualityChangeRandomMask_ArcFace} and \ref{fig:ImageQualityChangeRandomMask_FaceNet}, the distribution of image quality changes affected by the image degradation is shown.
The image quality change
\begin{align}
\Delta_{\hat{Q}} = \hat{Q}_{I_{org}} - \hat{Q}_{I_{mask}}
\end{align}
represents the difference between the FIQ of an unmodified image $I_{org}$ and a masked image $I_{mask}$.
A positive $\Delta_{\hat{Q}}$ indicates that the FIQ is successfully degraded in presence of the mask.
For the majority of the Random Mask dataset images, such positive values of $\Delta_{\hat{Q}}>0$ are observed for ArcFace.
For FaceNet, low mask sizes ($s=10, 20$) do not particularly affect the FIQ.
Only for medium or larger mask sizes ($s=30,40,50$) the FIQ is successfully degraded.

In Figures \ref{fig:PixelQualityChangeRandomMask_ArcFace} and \ref{fig:PixelQualityChangeRandomMask_FaceNet}, the distribution of the mean pixel-quality change in the masked area is shown.
The mean pixel-quality change 
\begin{align}
\Delta_p = \dfrac{1}{|\mathcal{P}|} \sum_{i,j \in \mathcal{P}} p_{i,j}^{org} - p_{i,j}^{mask}
\end{align}
measures the average difference in the pixel-level qualities $p_{i,j}$ of the unaltered and the masked images in the masked area $\mathcal{P}$.
Similar to the image quality change $\Delta_{\hat{Q}}$, a positive $\Delta_p$ indicates that the masks lead to degraded pixel qualities.
For ArcFace, a large portion of the distributions has positive values for all mask sizes.
For FaceNet, this behaviour is only observed for large mask sizes ($s\geq 30$) since the utilized model tends to be robust against smaller disturbances (see Figures \ref{fig:ImageQualityChangeRandomMask_FaceNet} and \ref{fig:PixelQualityChangeRandomMask_FaceNet}).
In general, the proposed methodology catches the added disturbances with both models and assigns them with significantly lower pixel-level qualities, demonstrating that the produced pixel-level qualities are meaningful.

\vspace{-2mm}
\paragraph{Qualitative Analysis -}

\begin{figure*}
\centering
\includegraphics[width=0.9\textwidth]{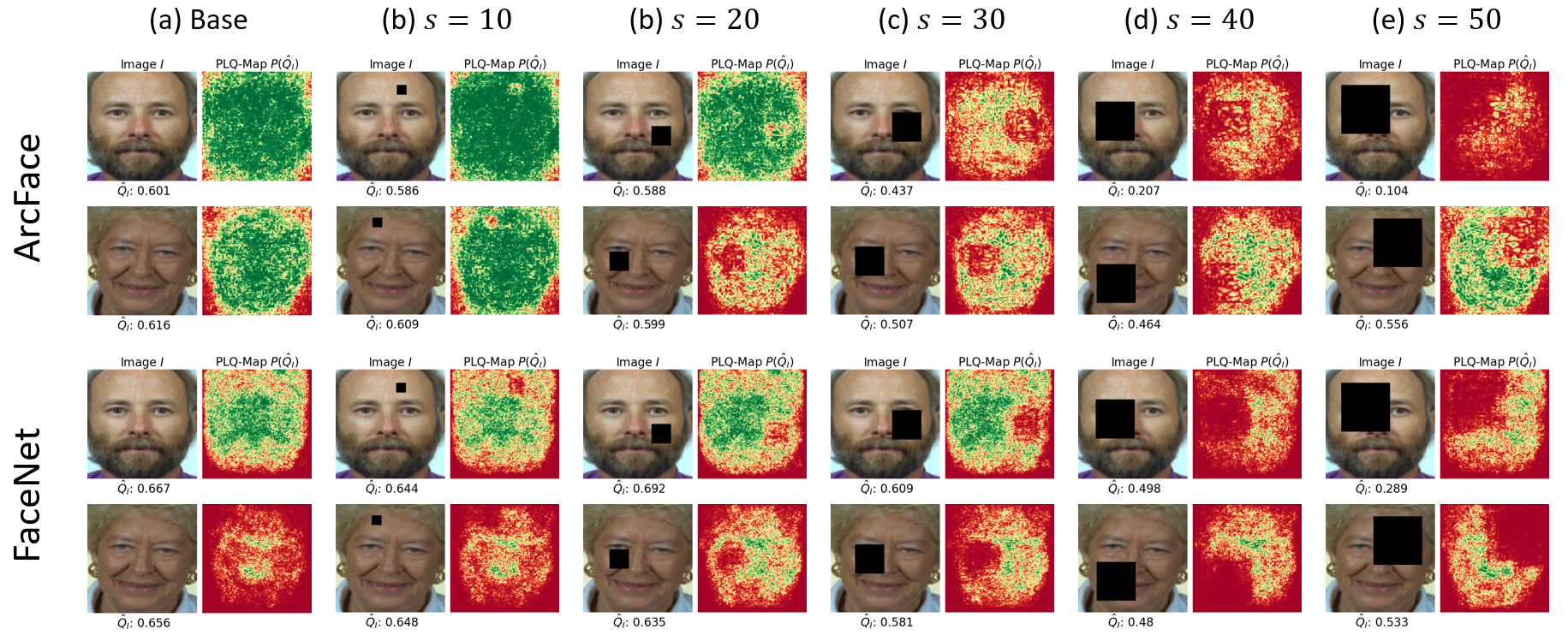}
\caption{\textbf{PLQ-explanation maps for random masks} - For two random identities, their masked and unmasked images are shown with their corresponding PLQ-maps. In general, the effect of the mask on the PLQ-map is clearly visible demonstrating the effectiveness of the proposed approach to detect disturbances.}
\label{fig:RandomMaskResults}
\end{figure*}

In Figure \ref{fig:RandomMaskResults}, the PLQ explanation maps for two random identities are shown over several mask sizes and locations for both FR models.
Moreover, the face images and their PLQ-maps are shown without masks.
For all mask sizes, the masked area is assigned with significantly smaller pixel-level quality values than the surrounding pixels. 
Consequently, the effect of the masks on the PLQ-map is clearly visible demonstrating that the proposed PLQ assessment approach can detect such disturbances.
Besides, it can be seen that the FIQ depends not only on the size but also on the position of the mask.

%

\subsection{PLQ-Maps on ICAO-Incompliant Images}
\label{sec:ICAO_Incompliance}

\begin{figure*}
\centering
\includegraphics[width=0.9\textwidth]{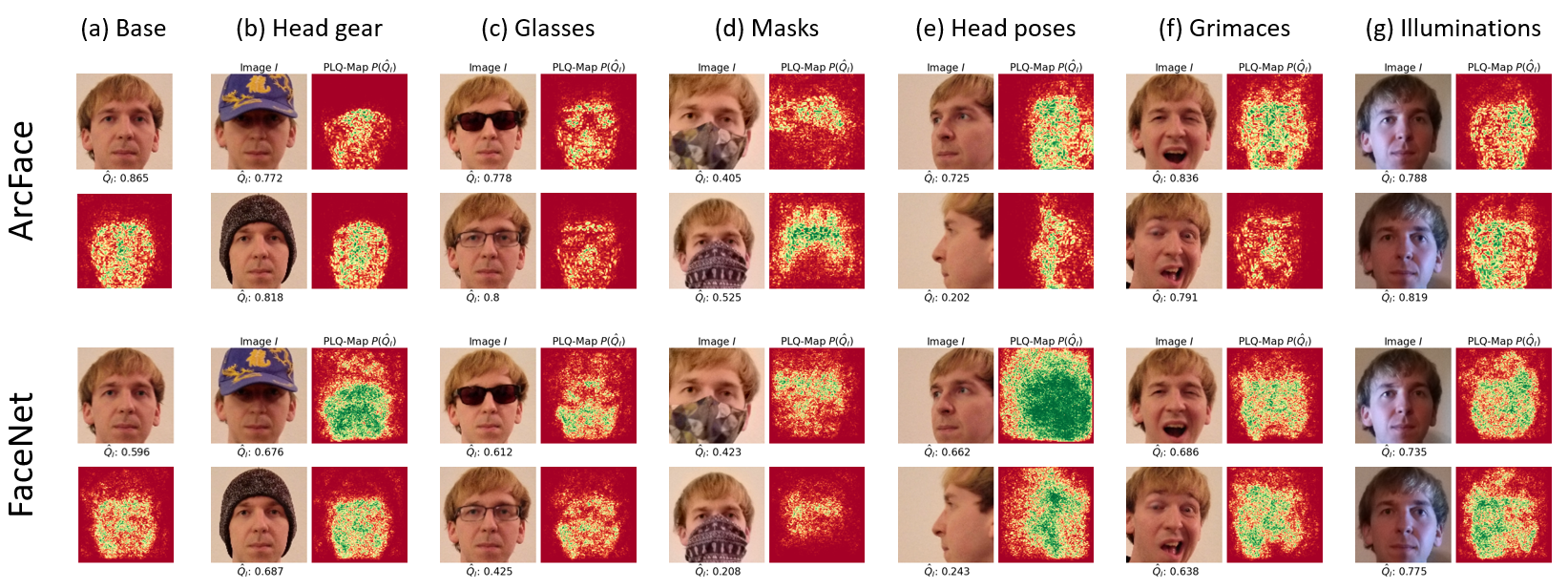}
\caption{\textbf{PLQ-explanation maps for ICAO imcompliant images} - One ICAO-compliant image and twelve images with imcompliances are shown with their corresponding PLQ-maps. Occlusions (b,c,d), distorted areas of the face (f), and reflections result in low-pixel qualities. \vspace{-2mm}}
\label{fig:ICAOResults}
\end{figure*}

Lastly, we analyse the proposed approach by investigating the effect of ICAO incompliances on the PLQ-maps.
In Figure \ref{fig:ICAOResults}, the ICAO-compliant and several face images with different violations of these specifications are shown with their corresponding PLQ-maps for both FR models.
The shown violations include wearing headgear, glasses, and masks, non-frontal head poses, non-neutral expressions, and irregular illuminations.
For the ICAO-compliant reference image (a), the area of the face is clearly visible and the PLQ-maps show high pixel-qualities in this area.
The same goes for wearing headgears (b) and masks (d) except that the occluded part of the face is assigned with low pixel-quality values.
For glasses (c), low-pixel qualities are assigned at the frame of the glasses while the darkened eyes are assigned with higher quality values.
Also for the non-frontal head poses (e), the distinction between background and face is clearly visible in the PLQ-maps.
Considering non-neutral expressions (f), areas that are distorted compared to neutral expressions are marked as low pixel-quality.
Lastly, an interesting effect is observed for non-uniform illuminations.
Not the illumined side of the face is assigned with higher quality, instead, the reflections might result in lower pixel-level qualities and the side away from the light is assigned with higher-quality values.
Generally, each of the ICAO incompliances can be interpreted with the PLQ maps.

\subsection{Summary}
\label{sec:Summary}

The investigations quantitatively and qualitatively demonstrated the effectiveness of the proposed PLQ assessment approach from two opposite directions and by comparing it with ICAO incompliances.
First, in images with low FIQ, low-pixel quality regions are detected and it was shown that inpainting these low-pixel quality regions lead to an increased FIQ.
Consequently, the assessment of low pixel-quality areas with the proposed method was correct.
Second, images with high FIQ were degraded by placing random masks on the images.
The PLQ-maps reliably assigned low-pixel qualities to the areas of disturbances demonstrating the effectiveness of the PLQ assessment.
Lastly, the usefulness of the method was proven by showing that the method can detect ICAO incompliances with the PLQ-maps.

Generally, it was shown that the FIQ of an image not only depends on the size of the disturbances (masks) but also on their location on the face.
The same applies to PLQ values.
The pixel-qualities do not contribute equally and independently to the overall FIQ.
Instead, the composition of pixel-qualities might have a stronger impact on the FIQ and thus, on the FR performance.
Consequently, a higher FIQ does not necessarily imply that the image must contain more  high-quality pixels.
Moreover, the PLQ-maps are dependent on the utilized FR models and how it is trained.
In our case, the face recognition performance, and thus the FIQ assessment performance, of ArcFace is higher than for FaceNet \cite{DBLP:conf/cvpr/TerhorstKDKK20}.
Consequently, the quality values and the PLQ-maps are more stable and precise on the ArcFace model.
Future works may analyse how well model biases (e.g. for demographics) might be reflected in the PLQ-maps due to the use of model-specific FIQ estimations \cite{DBLP:conf/icb/TerhorstKDKK20}.

%
%
%
%
%
%





\section{Limitations and Ethical Considerations}

When applying the proposed methodology we propose to use Gradient Clipping \cite{DBLP:conf/iclr/ZhangHSJ20} for the backpropagation of quality-based gradients.
This aims to avoid exploding gradients and thus, unreasonable PLQ-maps.
Details are provided in the supplementary material.
Moreover, we want to emphasizes that, depending on the application, inpainting should not carelessly be used to improve FIQ of face images since it might add artefacts leading to wrong matching decisions \cite{DBLP:conf/icb/MathaiMA19}.

\section{Conclusion}
\label{Conclusion}

The high performance of current FR systems is driven by the quality of its samples.
To ensure a high sample quality, for instance, in an automated border control scenario, the FIQ of a captured face is determined.
Consequently, a captured face might be rejected during enrolment without a hint of the quality-decreasing factor.
In this work, we proposed a methodology to compute pixel-level quality explanation maps to determine which regions of the face have a high and low utility for recognition.
Therefore, the proposed approach provides feedback on the utility of a face image that is understandable for humans.
Given an arbitrary FR network, we propose a training-free approach that determines the pixel-level quality maps for a face image in three steps.
In the first step, a model-specific quality estimate for the image is calculated, modified, and used, in the second step, to construct a quality regression model for the input image.
In the third step, quality-based gradients are back-propagated through the model and converted into pixel-level quality maps.
The experiments qualitatively and quantitatively demonstrated the effectiveness of the proposed approach in estimating pixel-level qualities.
This was shown on real and artificial disturbances and by comparing to ICAO-incompliant images.
Moreover, the experiments allowed us to gain more insights into the functionality of FR systems. 
For instance, it was shown that well-illuminated areas of the face get assigned with significantly lower pixel-qualities than the shaded area of the face.
Consequently, the shaded areas provide more important information for the FR models.
To summarize, the proposed approach can be applied to arbitrary FR networks, does not require training, and provides a pixel-level utility description of the input face that can be used to (a) deepen the understand of how face recognition systems work, (b) enhance the performance of these systems, and (c) to provide understandable feedback of why an image is accepted or rejected during enrolment due to quality concerns.

\paragraph{Acknowledgement}
This research work has been funded by the German Federal Ministry of Education and Research and the Hessen State Ministry for Higher Education, Research and the Arts within their joint support of the National Research Center for Applied Cybersecurity ATHENE.
Portions of the research in this paper use the FERET database of facial images collected under the FERET program, sponsored by the DOD Counterdrug Technology Development Program Office.
This work was carried out during the tenure of an ERCIM ’Alain Bensoussan‘ Fellowship Programme.


{\small
\bibliographystyle{ieee_fullname}
\bibliography{egbib}
}

\clearpage
\newpage

\section{Supplementary}

To provide more comprehensive insights into the proposed methodology, experiments, and results, in the following, more details on these points are given.
In Section \ref{sec:Sup_Databases}, additional information on the used databases is given.
This includes sample images of the databases as well as details on their creating processes.
In Section \ref{sec:Sup_FIQscaling}, provides additional information on the scaling of the FIQ values.
Section \ref{sec:Sup_ExplodingGradients} investigates the phenomenon of exploding gradients in the proposed methodology and how to deal with it.
Finally, more comprehensive results are shown in Section \ref{sec:Sup_MoreResults} to investigate the effect of coloured disturbances on the PLQ maps and to analyse the PLQ maps on all ICAO-incompliant face images.

\subsection{Additional Information on the Used Databases}
\label{sec:Sup_Databases}
In this section, we provide additional information on the utilized databases.
We show more example images of the used databases and discuss the licenses and capturing processes of these database.
This aims at enhancing the understanding of the databases and their creation process.

\begin{figure*}
\centering
\includegraphics[width=0.8\textwidth]{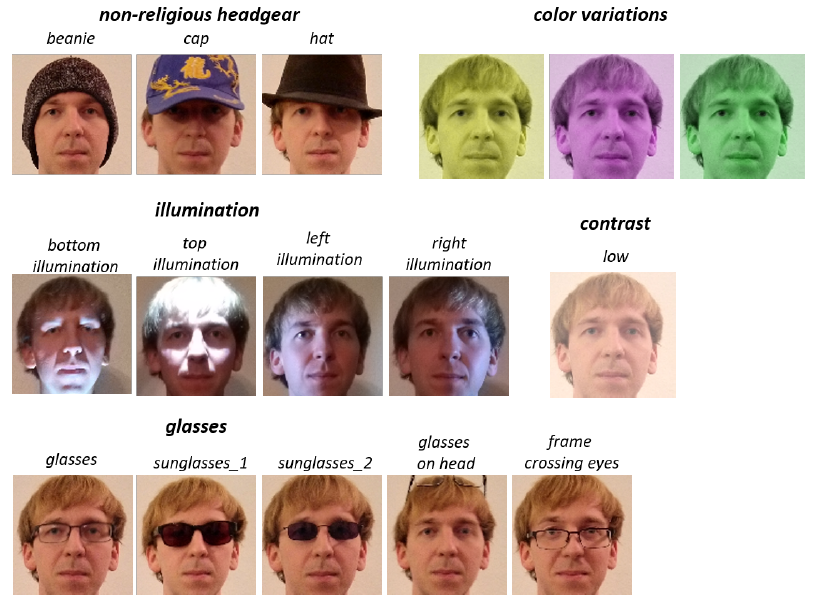}
\includegraphics[width=0.8\textwidth]{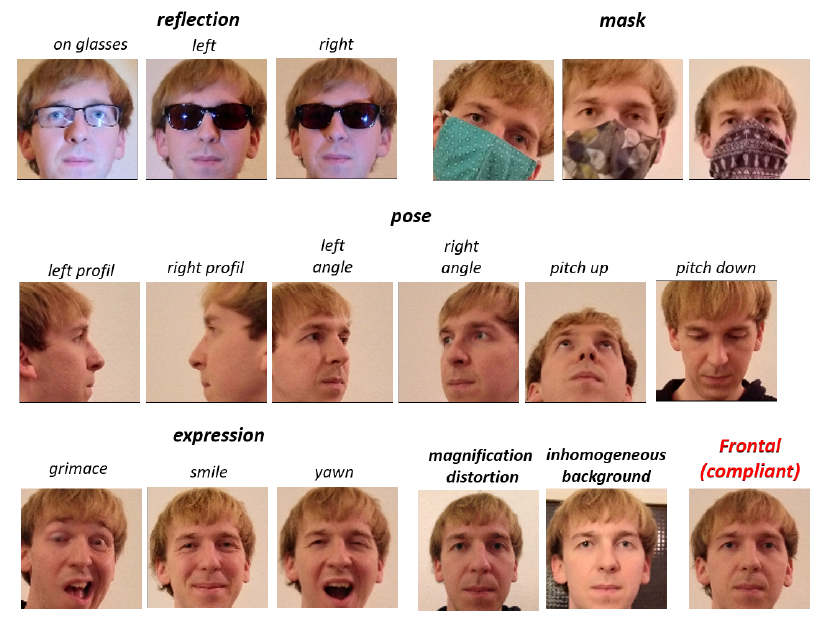}
\caption{\textbf{Inhouse ICAO Incompliance dataset} - Images created to violate the ICAO specifications \cite{ICAO9303}. This includes: incompliant headgear, color variations, illumination, blur, wearing glasses, low contrast, reflection, wearing masks, pose, expressions, magnification distortion and inhomogeneous background.}
\label{fig:ICAO_samples}
\end{figure*}

In Figure \ref{fig:ICAO_samples}, the images of our Inhouse ICAO Incompliance dataset are shown.
The images show various capture violations regarding the International Civil Aviation Organization as shown in the figure.
The dataset is private and was only collected for the purpose of this research.
Using this data for other research projects requires a confirmation of the subject first.
Before capturing the images of the dataset, the subject's consent was explicitly given.

\begin{figure*}
\centering
\includegraphics[width=0.6\textwidth]{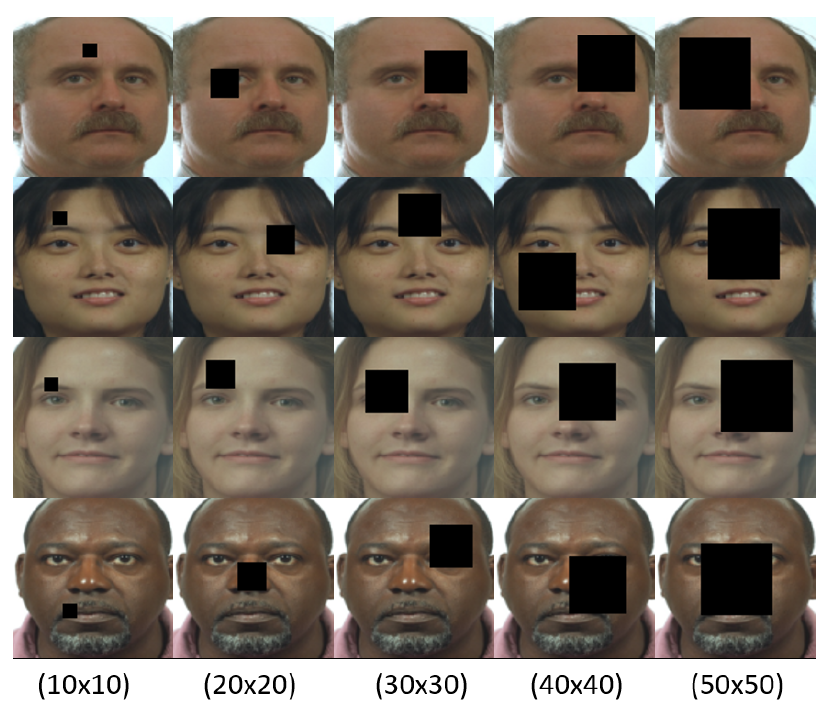}
\caption{\textbf{Samples of the Random Mask dataset} - Masks of sizes ($10 \times 10$) to ($50 \times 50$) pixels are randomly placed on the images of ColorFeret \cite{ColorFERET}. The dataset is used to perform the quantitative analysis as described in Section \ref{sec:Investigations}. More information on this database was given in Section \ref{sec:Databases}}.
\label{fig:RandomMask_samples}
\end{figure*}

In Figure \ref{fig:RandomMask_samples}, sample images of two random identities for the Random Mask dataset are shown.
As described in Section \ref{sec:Databases}, the images were created by subsequentially placing five black square-shaped masks of various sizes on a face image with random locations and make use of ColorFeret \cite{ColorFERET} images.
The ColorFeret \cite{ColorFERET} database is restricted to face recognition research.
During the data collection, the different subjects were photographed in 15 session over three years under controlled conditions.
Detailed license information can be found under \url{https://www.nist.gov/system/files/documents/2019/11/25/colorferet_release_agreement.pdf}.
More details can be found in \cite{ColorFERET} and \url{https://www.nist.gov/itl/products-and-services/color-feret-database}.

\begin{figure*}
\centering
\includegraphics[width=0.9\textwidth]{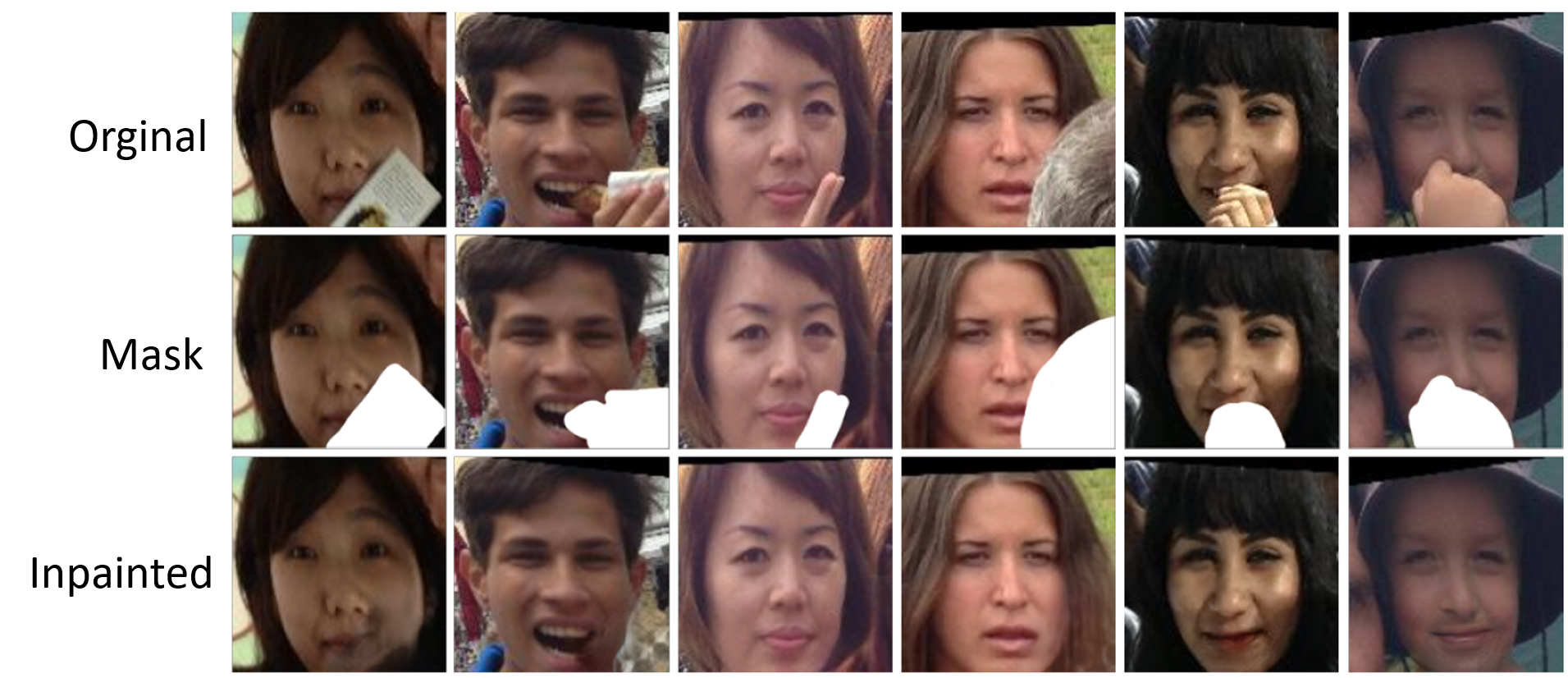}
\caption{\textbf{Samples of the Pre- and Inpainted dataset} - The top row shows some unmodified samples used for the dataset. The middle row presents the mask areas for the samples and the last row shows the resulting inpainted images. }
\label{fig:Inpainting_samples}
\end{figure*}

Finally, Figure \ref{fig:Inpainting_samples} shows samples of the Pre- and Inpainted dataset. 
In the top row, the original face images with various quality-decreasing factors (here occlusions) are shown. 
The middle row shows the masks (white) drawn by a student after receiving feedback on the low-quality regions via the PLQ explanation maps.
Finally, the bottom row shows the resulting image when masked faces are given to an inpainting algorithm.
The images of the Pre- and Inpainted dataset come from the VGGFace2 \cite{DBLP:conf/fgr/CaoSXPZ18} and the Adience \cite{Eidinger:2014:AGE:2771306.2772049} dataset and consists of individual images which were uploaded to the internet and tagged as publicly available by the original author.
More information can be found in \cite{DBLP:conf/fgr/CaoSXPZ18} and \cite{Eidinger:2014:AGE:2771306.2772049} and under \url{http://www.robots.ox.ac.uk/~vgg/data/vgg_face2} and \url{https://talhassner.github.io/home/projects/Adience/LICENSE.txt}.

\subsection{Additional Information on the FIQ Scaling}
\label{sec:Sup_FIQscaling}

\begin{figure*}
\centering
	
\subfloat[Unscale FIQ distribution - ArcFace \label{fig:UnscaleFIQDistribution_ArcFace}]{%
	\includegraphics[width=0.45\textwidth]{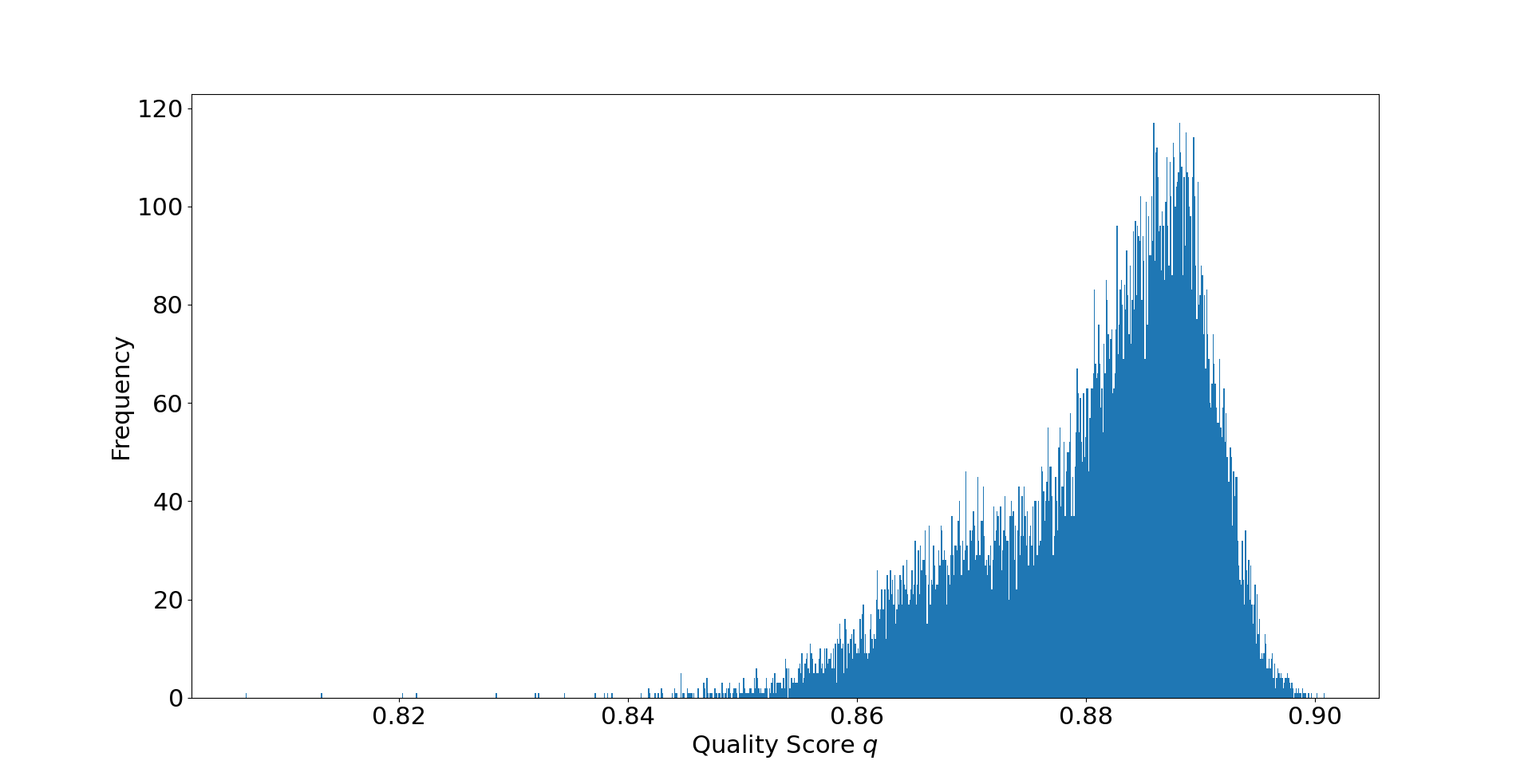}} 	
\subfloat[Scaled FIQ distribution - ArcFace \label{fig:ScaleFIQDistribution_ArcFace}]{%
	\includegraphics[width=0.45\textwidth]{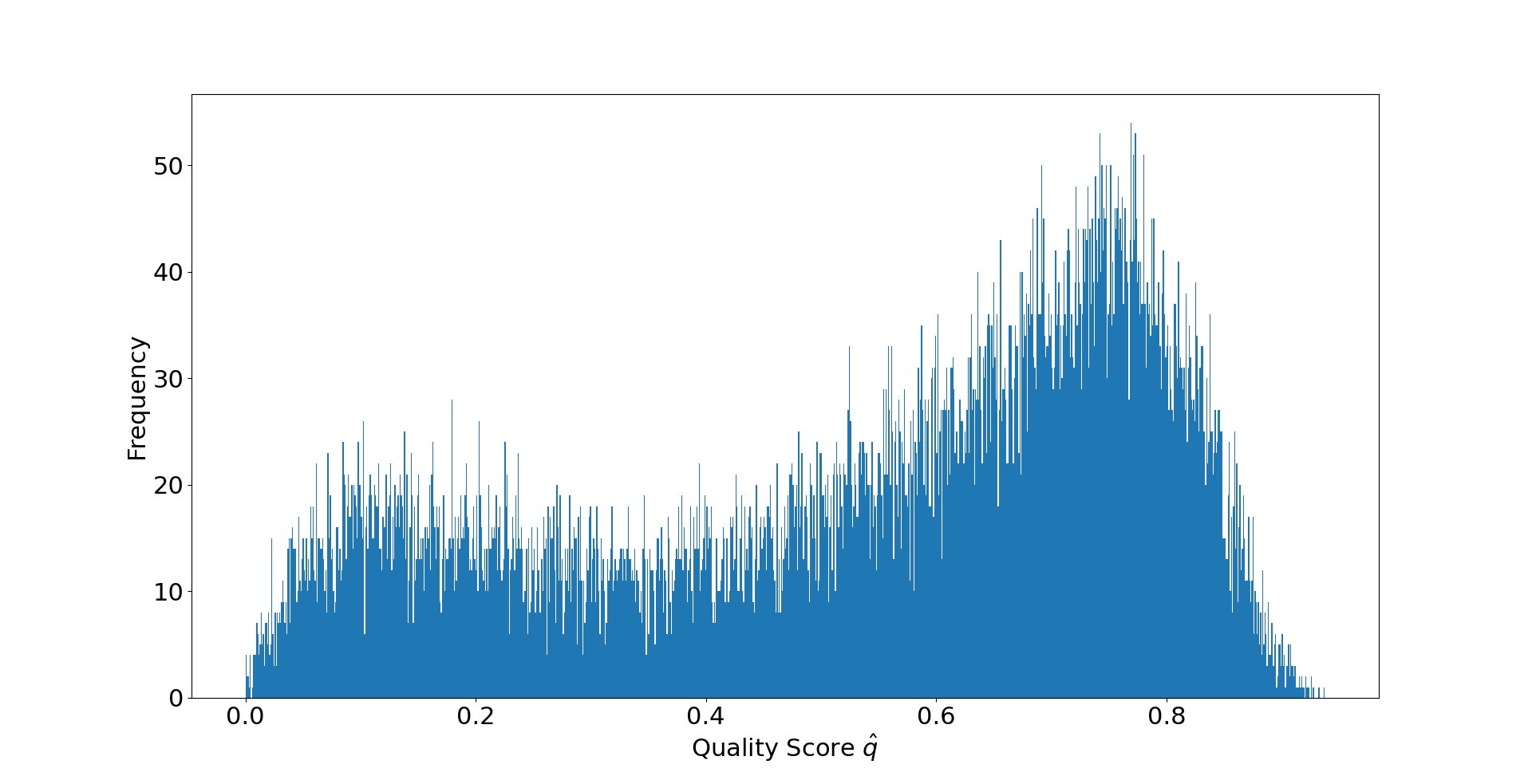}}

\subfloat[Unscale FIQ distribution - FaceNet \label{fig:UnscaleFIQDistribution_FaceNet}]{%
	\includegraphics[width=0.45\textwidth]{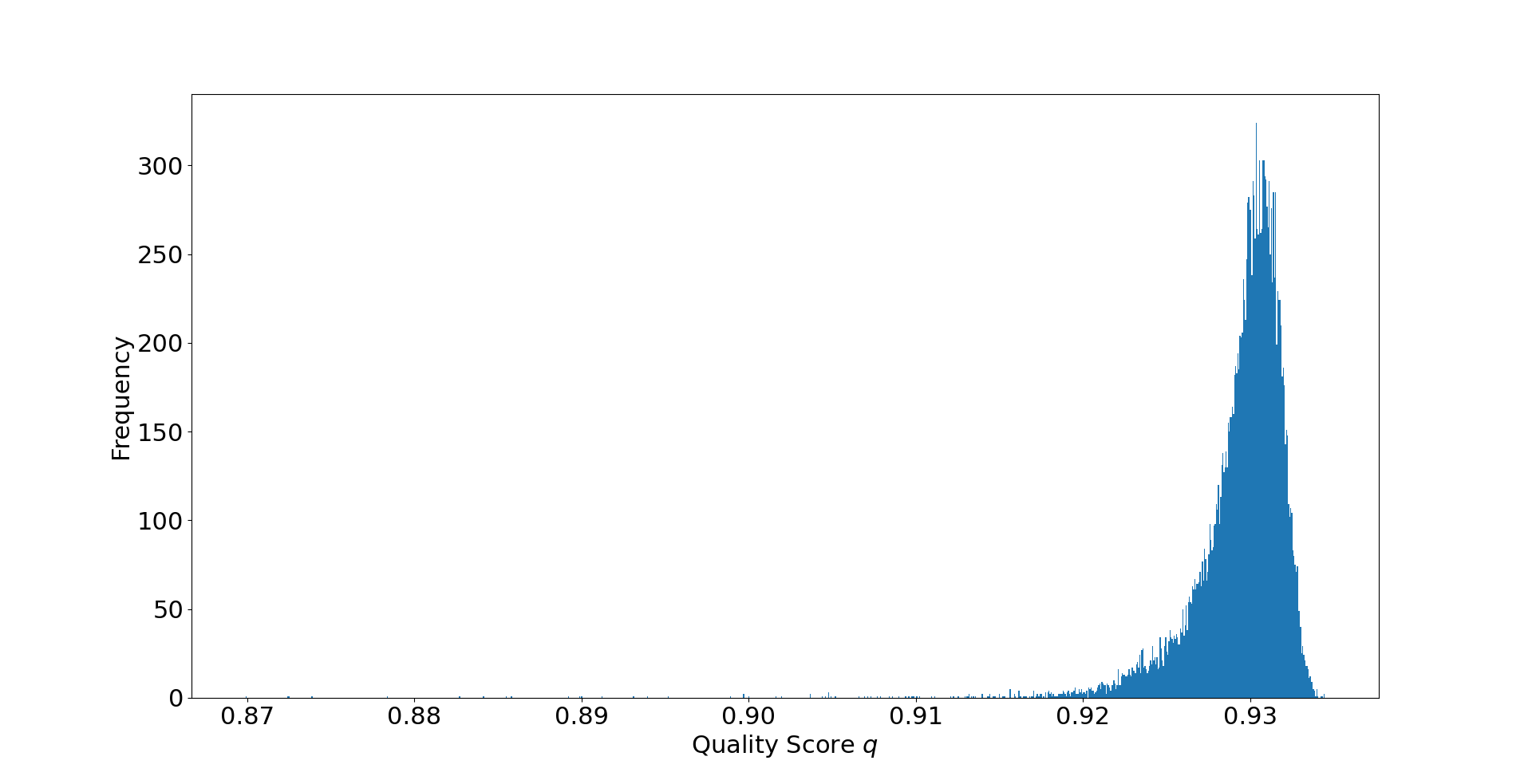}} 	
\subfloat[Scaled FIQ distribution - FaceNet \label{fig:ScaleFIQDistribution_FaceNet}]{%
	\includegraphics[width=0.45\textwidth]{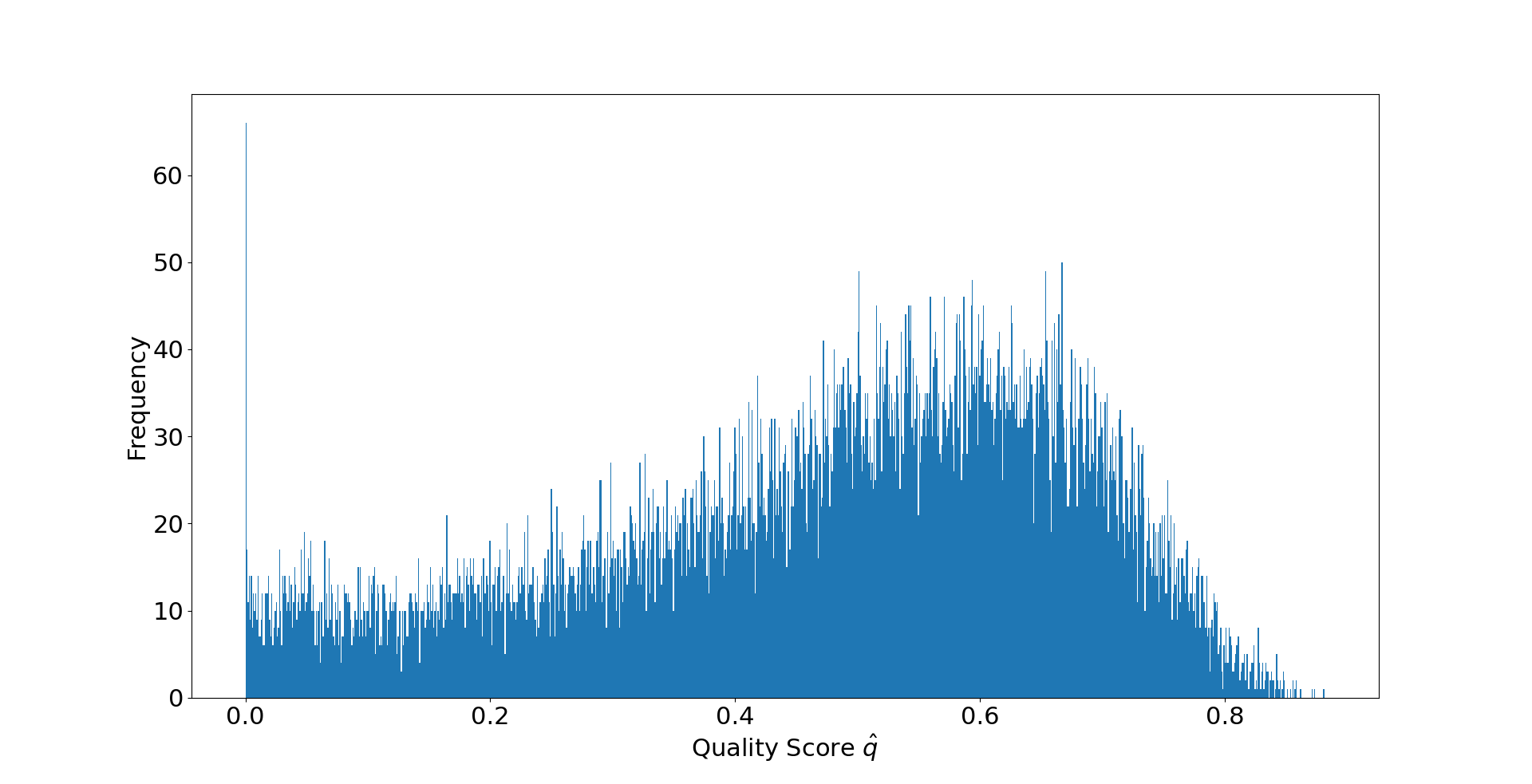}}
	
\caption{\textbf{FIQ distributions on Adience} - The FIQ distributions of ArcFace (a,b) and FaceNet (c,d) are shown before and after the quality scaling explained in Equation \ref{eq:QualityScaling}. While the unscaled quality values are in a narrow range, the scaled qualities are in a more convenient range of [0,1] that makes it easier to understand and interpret.
}
\label{fig:FIQ_Scaling}
\end{figure*}

In Section \ref{sec:Methodology}, we mentioned that the FIQ values are dependent on the utilized model and often appear in a narrow range. 
Figures \ref{fig:UnscaleFIQDistribution_ArcFace} and \ref{fig:UnscaleFIQDistribution_FaceNet} show the distributions of the FIQs on the Adience dataset. 
Although only the order of the quality values is crucial for FIQA, a narrow range makes it harder for humans to compare.
To make it easier comparable, we applied the scaling function shown in Equation \ref{eq:QualityScaling} to bring the values in a range of [0,1] as shown in Figure \ref{fig:FIQ_Scaling}.

\subsection{The Problem of Exploding Gradients}
\label{sec:Sup_ExplodingGradients}

\begin{figure*}
\centering
\subfloat[ArcFace - 1\label{fig:ExplodingGradient_1_ArcFace}]{%
	\includegraphics[width=0.245\textwidth]{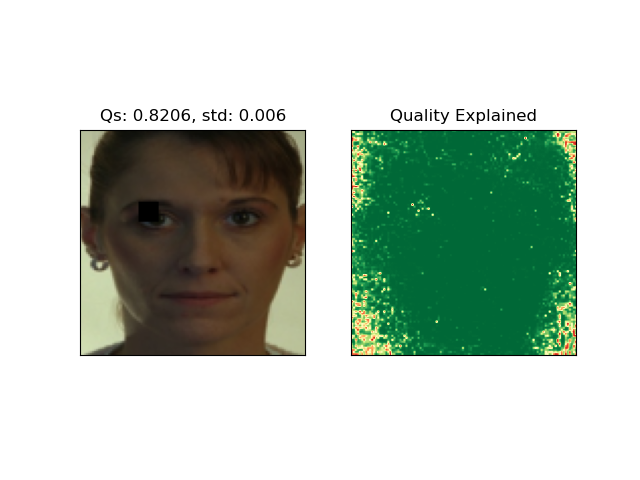}}
\subfloat[ArcFace - 2\label{fig:ExplodingGradient_2_ArcFace}]{%
	\includegraphics[width=0.245\textwidth]{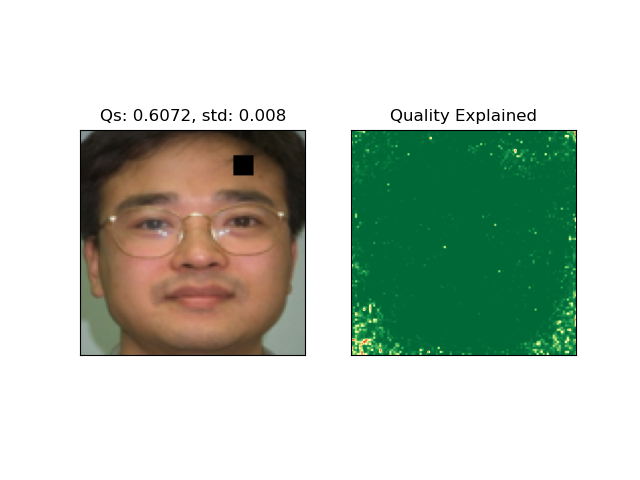}} 
\subfloat[FaceNet - 1\label{fig:ExplodingGradient_1_FaceNet}]{%
	\includegraphics[width=0.245\textwidth]{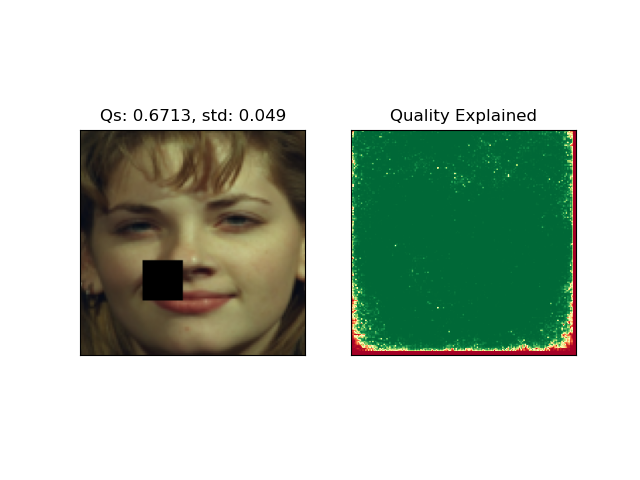}}
\subfloat[FaceNet - 2\label{fig:ExplodingGradient_2_FaceNet}]{%
	\includegraphics[width=0.245\textwidth]{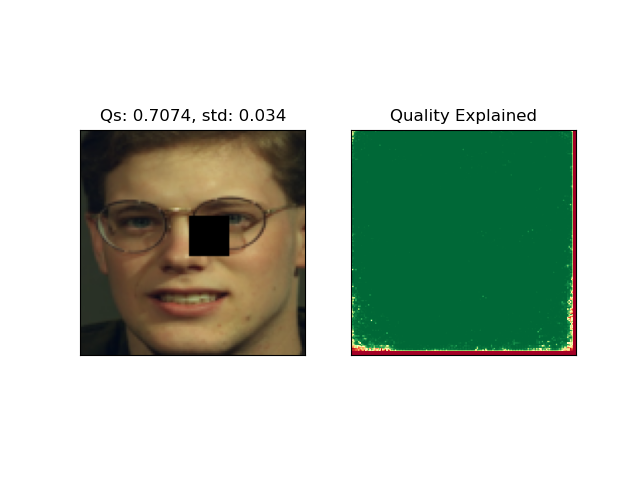}}		
\caption{\textbf{Samples of failed PLQ maps} - For some sample images extreme PLQ maps appear due to exploding gradients during the backpropagation step. Simple Gradient Clipping \cite{DBLP:conf/iclr/ZhangHSJ20} solves the problem.}
\label{fig:ExplodingGradients}
\end{figure*}

During the quantitative experiments, we noticed that some images show extreme PLQ-Maps with respect to other PLQ-maps.
Some examples are shown in Figure \ref{fig:ExplodingGradients}.
This happens due to exploding gradients during the backpropagation of the quality-based gradients.
Even though this only happens rarely, a simple gradient clipping \cite{DBLP:conf/iclr/ZhangHSJ20} solves this problem.

\subsection{More Comprehensive Results}
\label{sec:Sup_MoreResults}

Due to the restricted page limit the main paper focus on the most essential results that leads to a clear message of the paper.
More comprehensive results are presented in the following.
First, the impact of disturbances of different colours on the PLQ maps is studied.
Second, the results of all ICAO-incompliant images are presented and analysed.

\subsubsection{Color Study}

\begin{figure*}
\centering
	
\subfloat[ArcFace \label{fig:ColorStudy_ArcFace}]{%
	\includegraphics[width=0.45\textwidth]{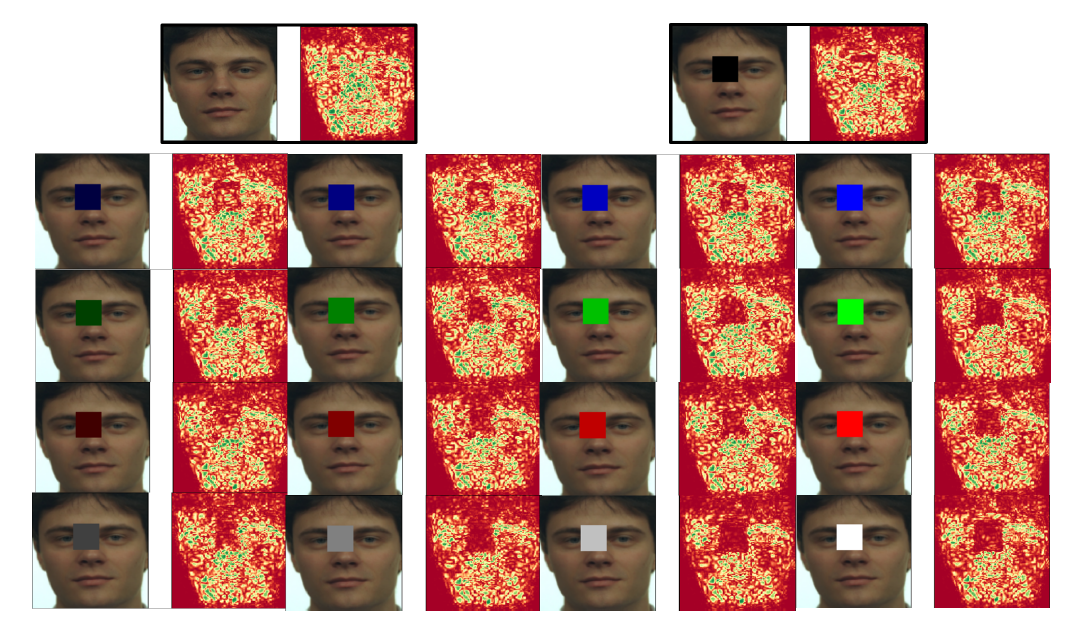}} 	
\subfloat[FaceNet \label{fig:ColorStudy_FaceNet}]{%
	\includegraphics[width=0.45\textwidth]{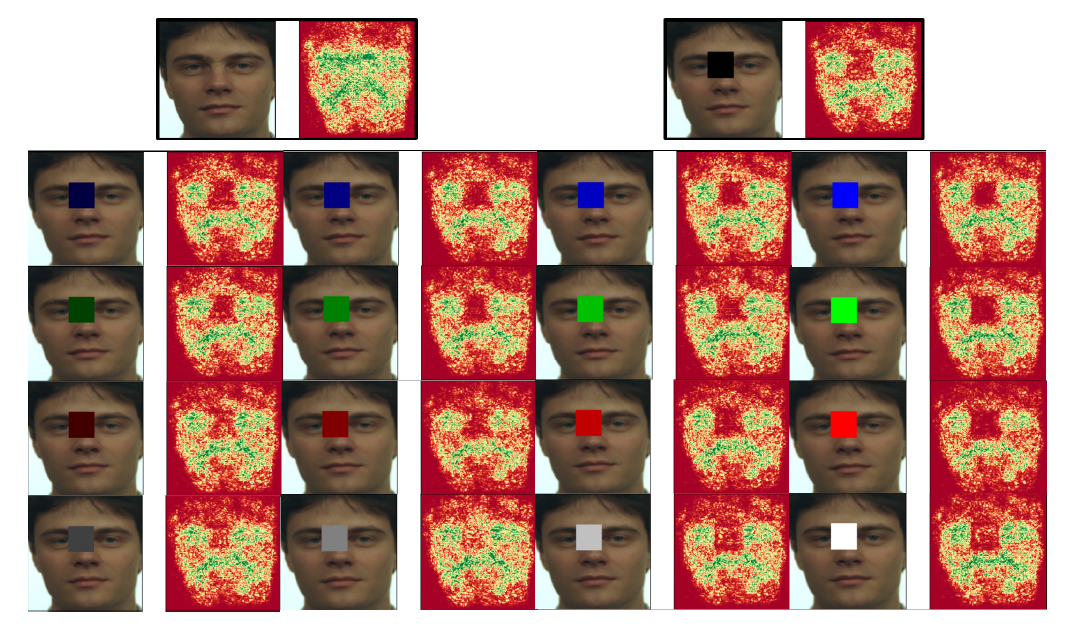}} 	
\caption{\textbf{Analysis of the impact on coloured disturbances on PLQ maps} - In (a), the PLQ maps for the ArcFace model are shown. In (b), the same is shown for the FaceNet model. The results show that the PLQ maps are generally quite robust to disturbances of different colors.
}
\label{fig:ColorStudy}
\end{figure*}

To analysis the impact of disturbances of different colours on the PLQ maps, Figure \ref{fig:ColorStudy} shows these maps for masks of 12 different colors.
In nearly all cases, the masks can be easily detected on the PLQ maps.
Only disturbances that have a similar color to the subject's skin are harder to detect.
In general, disturbances of different colors do not affect the PLQ maps.

\subsubsection{Full Results on the ICAO-Incompliance Dataset}

In Section \ref{sec:ICAO_Incompliance}, the PLQ maps were analysed on 13 prototype images. 
These images show different ICAO incompliances such as wearing head gear, glasses, masks, showing different head poses, grimaces and different illumination settings.
In this section, also the remaining images, and the corresponding PLQ maps, are shown and analysed.

\begin{figure*}
\centering
\subfloat[Headgear \label{fig:ICAO_results_headgear}]{%
	\includegraphics[width=0.5\textwidth]{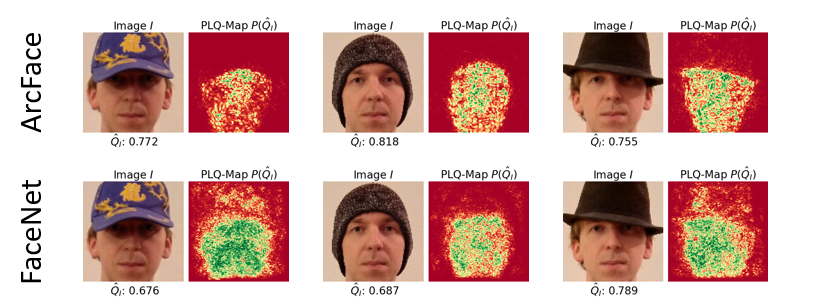}}
\subfloat[Color variations \label{fig:ICAO_results_colorvariants}]{%
	\includegraphics[width=0.5\textwidth]{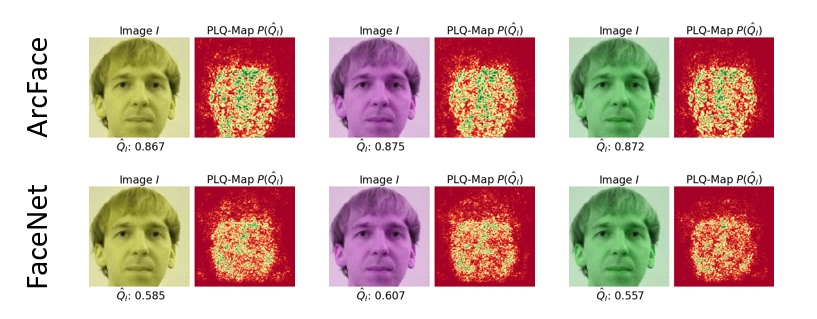}}
	
\subfloat[Illumination \label{fig:ICAO_results_illumination}]{%
	\includegraphics[width=0.7\textwidth]{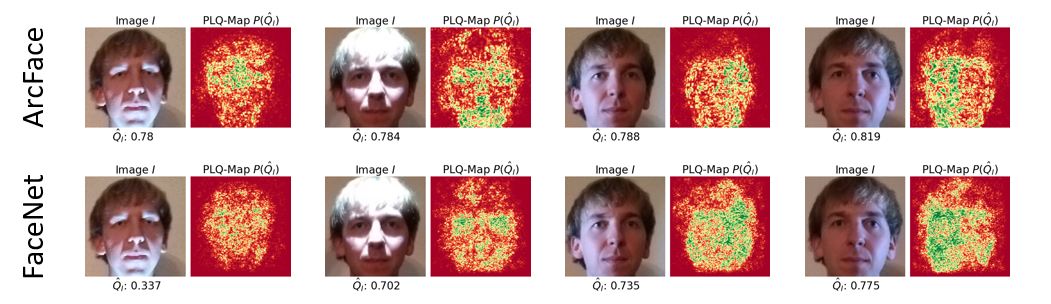}}
	
\subfloat[Glasses \label{fig:ICAO_results_glasses}]{%
	\includegraphics[width=0.8\textwidth]{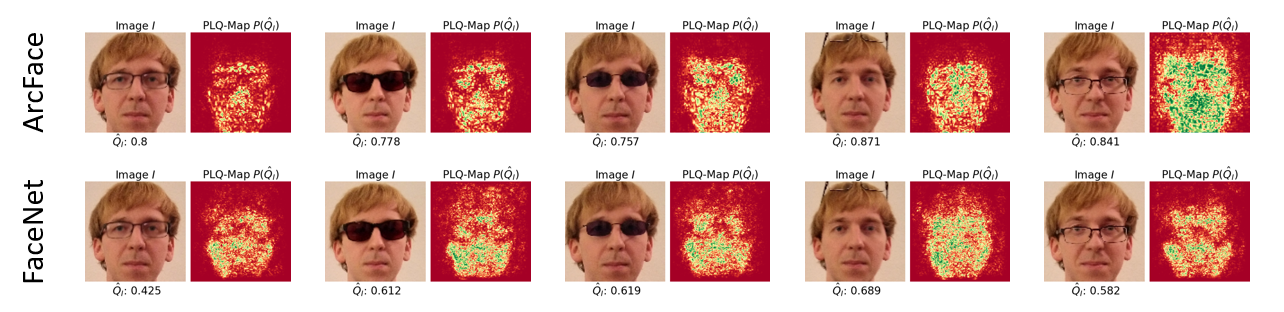}}
	
\caption{\textbf{PLQ-Maps for ICAO incompliances Part 1} - Shown are the PLQ-Maps for ICAO incompliant face images through hear gear, color variations, illuminations, and glasses. The proposed method turns out to be robust against color variation (Figure \ref{fig:ICAO_results_colorvariants}). Moreover, it steadily assigns occlusions of the face as low-quality areas (Figures \ref{fig:ICAO_results_headgear}, \ref{fig:ICAO_results_glasses}). Counterintuitively, less-illuminated areas of the face tend to have a higher importance for the FR models than areas with a strong illumination (Figure \ref{fig:ICAO_results_illumination}).}
\label{fig:ICAO_results_1}
\end{figure*}

In Figure \ref{fig:ICAO_results_1}, the PLQ maps for four ICAO incompliances are shown.
Figure \ref{fig:ICAO_results_headgear} shows the effect of wearing head gears. 
For both FR models, the regions covered by the head gears are assigned with lower PLQ values.
In Figure \ref{fig:ICAO_results_colorvariants}, the effect of coloured images on the PLQ maps is analysed.
It turns out that the proposed method is robust against these color variations.
Figure \ref{fig:ICAO_results_illumination} analyses the effect of different illumination settings.
Counterintuitively, it shows that not the well-illuminated side of the face is assigned with high quality values. 
Instead, the reflections lead to low pixel-quality areas and the side away from the light is assigned with higher-quality values.
Since the PLQ-maps are only based on the input image and the FR model, it shows that recognition with these systems works differently than from humans.
Illumination might even hurt the recognition performance of an FR system.
In Figure \ref{fig:ICAO_results_glasses}, the effect of different glasses and glass positions are analysed.
Independently on the used glasses, only the border of the glasses resulted in low-quality areas.

\begin{figure*}
\centering
\subfloat[Reflections \label{fig:ICAO_results_reflections}]{%
	\includegraphics[width=0.5\textwidth]{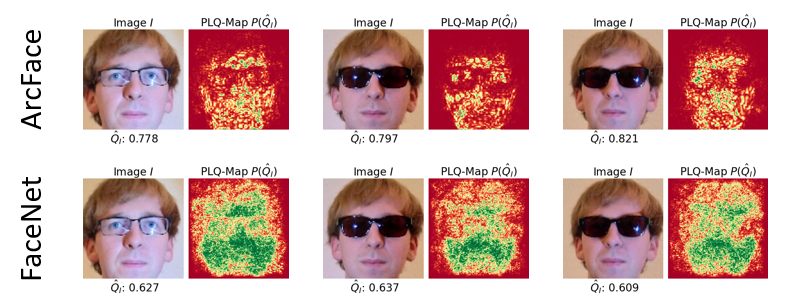}}
\subfloat[Masks \label{fig:ICAO_results_masks}]{%
	\includegraphics[width=0.5\textwidth]{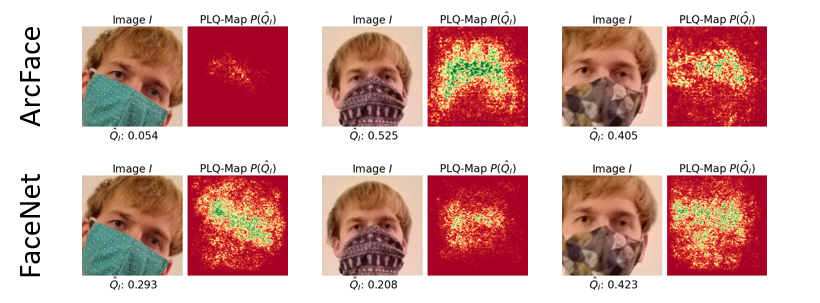}}
	
\subfloat[Pose \label{fig:ICAO_results_pose}]{%
	\includegraphics[width=0.8\textwidth]{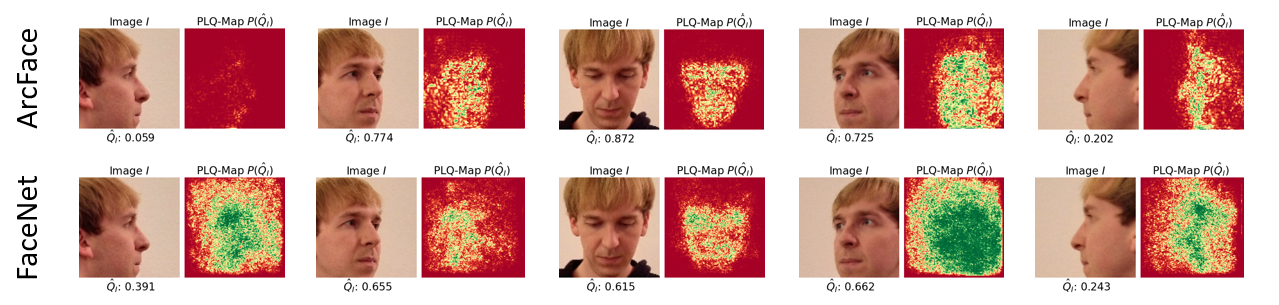}}
	
\subfloat[Expression \label{fig:ICAO_results_expression}]{%
	\includegraphics[width=0.5\textwidth]{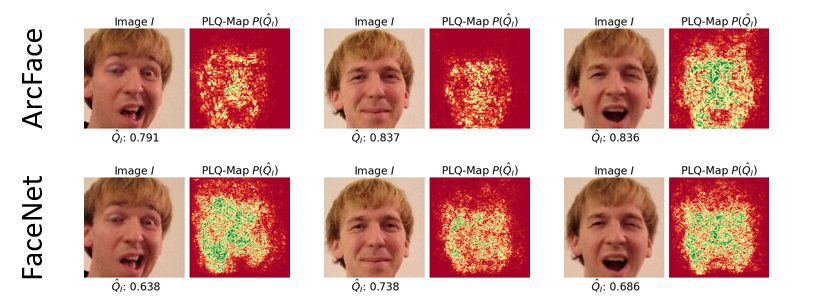}}
\subfloat[Others \label{fig:ICAO_results_others}]{%
	\includegraphics[width=0.5\textwidth]{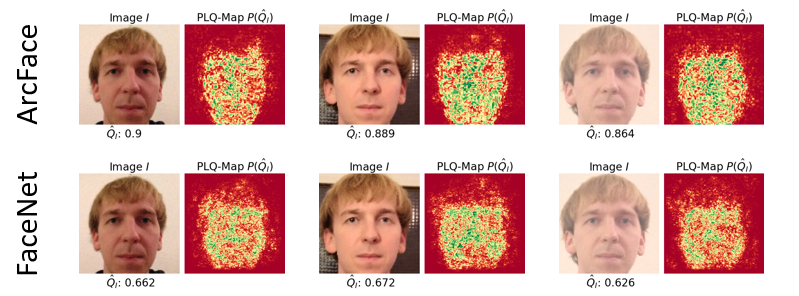}}
	
\caption{\textbf{PLQ-Maps for ICO incompliances - Part 2} - Shown are the PLQ-Maps for ICAO incompliant face images through reflections, masks, poses, expressions, magnification distortion, inhomogeneous background, and low contrast. The PLQ-Maps are robust against distortions, background, and low contrast (Figure \ref{fig:ICAO_results_others}) and are reliable able to assign face masks as low-quality regions (Figure \ref{fig:ICAO_results_masks}). Moreover, distorted areas due to grimaces are captured as low pixel quality and thus, are correctly assigned as low utility for recognition. }
\label{fig:ICAO_results_2}
\end{figure*}

In Figure \ref{fig:ICAO_results_2}, the results of the remaining ICAO incompliances are shown.
Figure \ref{fig:ICAO_results_reflections} analyses the effect of reflections on glasses.
It shows that such reflections decrease the PLQ values in the area of the reflection.
In Figure \ref{fig:ICAO_results_masks}, the effect of masks on the PLQ maps are shown.
The proposed methodology is steadily assigning low pixel-quality values to the masked area.
Figure \ref{fig:ICAO_results_pose} shows the effect of different head poses.
While the FaceNet model results in some inaccuracies when facing profile head poses, the PLQ maps of the ArcFace model captures the face regions well independently of the poses.
In Figure \ref{fig:ICAO_results_expression}, different facial expressions (grimaces) are investigated.
Interestingly, face areas that are distorted by the grimaces are assigned with lower pixel qualities, while unaffected areas are of higher quality.
Lastly, Figure \ref{fig:ICAO_results_others} analyses the effect of magnification distortion, inhomogeneous background and low contrast images.
In all three cases, the effect of these incompliances is minor demonstrating that the PLQ maps, as well as the underlying FR systems, are robust against these effects.

\end{document}